\PassOptionsToPackage{table}{xcolor}

\documentclass[sigconf,nonacm]{acmart}
\setcopyright{none}   %
\usepackage{graphicx}
\usepackage{xurl}
\usepackage{booktabs}
\usepackage{multirow}
\usepackage{tabularx}
\usepackage{bm}
\usepackage{amsmath}

\usepackage{amssymb}
\usepackage{xspace}
\usepackage{tikz}
\usepackage{pgfplots}
\pgfplotsset{compat=1.18}
\usepgfplotslibrary{fillbetween,groupplots}
\usetikzlibrary{positioning,arrows.meta,shapes.geometric,fit,backgrounds,calc,decorations.pathreplacing,spy}
\usepackage{algorithm}
\usepackage{algorithmic}
\usepackage[capitalise,noabbrev]{cleveref}

\newcommand{\etal}{\emph{et~al.}\xspace}
\makeatletter
\let\laplace\@undefined
\makeatother

\def\1{\bm{1}}

\def\vtheta{{\bm{\theta}}}

\def\vc{{\bm{c}}}
\def\vd{{\bm{d}}}

\def\vo{{\bm{o}}}

\def\vr{{\bm{r}}}

\def\vx{{\bm{x}}}

\DeclareMathAlphabet{\mathsfit}{\encodingdefault}{\sfdefault}{m}{sl}
\SetMathAlphabet{\mathsfit}{bold}{\encodingdefault}{\sfdefault}{bx}{n}

\def\sR{{\mathbb{R}}}
\def\sS{{\mathbb{S}}}

\newcommand{\laplace}{\mathrm{Laplace}} %

\newcommand{\E}{\mathbb{E}}
\newcommand{\Ls}{\mathcal{L}}

\newcommand{\lr}{\alpha}
\newcommand{\reg}{\lambda}

\newcommand{\sg}{\operatorname{sg}}

\graphicspath{{images/}}

\definecolor{methodNeRF}{RGB}{230,159,0}
\definecolor{methodFree}{RGB}{0,114,178}
\definecolor{methodOurs}{RGB}{0,158,115}
\definecolor{oiVermillion}{RGB}{213,94,0}
\definecolor{oiSkyBlue}{RGB}{86,180,233}
\definecolor{oiReddishPurple}{RGB}{204,121,167}
\definecolor{oiYellow}{RGB}{240,228,66}
\colorlet{renderLoss}{oiVermillion}
\colorlet{driftGreen}{methodOurs!70!black}
\colorlet{viewA}{oiVermillion}
\colorlet{viewB}{oiSkyBlue}
\colorlet{sceneViolet}{oiReddishPurple}

\pgfplotsset{
  method groupplot/.style={
    width=0.48\textwidth, height=4.2cm,
    every axis plot/.append style={semithick},
    tick label style={font=\scriptsize},
    label style={font=\small},
    title style={font=\small\bfseries},
    grid=major, grid style={gray!8},
    clip=false,
  },
  plot/nerf/.style={const plot, draw=methodNeRF, densely dashed,
                    fill=methodNeRF!25, fill opacity=0.4},
  plot/freenerf/.style={const plot, draw=methodFree, densely dotted,
                        fill=methodFree!25, fill opacity=0.4},
  plot/ours/.style={const plot, draw=methodOurs, solid,
                    fill=methodOurs!18, fill opacity=0.55},
  band/nerf/.style={methodNeRF!15, opacity=0.4},
  band/freenerf/.style={methodFree!15, opacity=0.4},
  band/ours/.style={methodOurs!12, opacity=0.4},
  line/nerf/.style={methodNeRF, densely dashed},
  line/freenerf/.style={methodFree, densely dotted},
  line/ours/.style={methodOurs, solid},
}

\newcommand{\legendOurs}{\textcolor{methodOurs}{Green\,(solid)}: DriftNeRF}

\definecolor{tabfirst}{rgb}{1,0.7,0.7}
\definecolor{tabsecond}{rgb}{1,0.85,0.7}
\definecolor{tabthird}{rgb}{1,1,0.7}
\definecolor{baselineGray}{gray}{0.92}
\newcommand{\first}[1]{\cellcolor{tabfirst}{#1}}
\newcommand{\second}[1]{\cellcolor{tabsecond}{#1}}
\newcommand{\third}[1]{\cellcolor{tabthird}{#1}}

\newcommand{\NotEval}{\textcolor{gray}{--}\xspace}

\title{DriftNeRF: Cross-Ray Density Drifting for Sparse-View Neural Radiance Fields}

\author{Juntang Wang}
\orcid{0009-0008-9812-7451}
\affiliation{%
  \institution{Harvard University}
  \city{Cambridge}\country{USA}}

\author{Yihan Wang}
\orcid{0009-0000-3586-9046}
\affiliation{%
  \institution{Duke University}
  \city{Durham}\country{USA}}

\author{Sheng'en Li}
\affiliation{%
  \institution{Duke Kunshan University}
  \city{Kunshan}\country{China}}

\author{Hao Wu}
\orcid{0009-0009-5792-3790}
\affiliation{%
  \institution{Duke Kunshan University}
  \city{Kunshan}\country{China}}

\author{Dongmian Zou}
\orcid{0000-0002-5618-5791}
\affiliation{%
  \institution{Duke Kunshan University}
  \city{Kunshan}\country{China}}

\author{Shixin Xu}
\authornote{Corresponding author.}
\orcid{0000-0002-8207-7313}
\affiliation{%
  \institution{Duke Kunshan University}
  \city{Kunshan}\country{China}}

\renewcommand{\shortauthors}{Wang et al.}

\begin{abstract}
Neural Radiance Fields (NeRF) synthesize photorealistic novel views from dense image collections, but degrade sharply when few views are available.
We attribute a specific, measurable defect to the objective itself: the rendering loss is a sum over independent rays and contains no term comparing densities that different rays assign to the same region of space, so nearby 3D points can settle on contradictory densities while every training pixel is reproduced correctly.
We make this quantity explicit by measuring cross-ray density disagreement directly during training, and show that it stays high in vanilla NeRF, falls only incidentally under frequency-based regularization, and moves together with held-out quality in the sparse regime.
We then close the gap with \textbf{DriftNeRF}, a drop-in regularizer that instantiates the drifting-field template on density samples: a masked exponential kernel aggregates densities from nearby points on other rays into a stop-gradient consensus target, and the resulting drifting field penalizes unsupported disagreement while leaving agreed-upon boundaries intact.
The regularizer reads only sampled positions, ray identifiers, and predicted densities, so it requires no pretrained model, external geometry, or additional images, and transfers across field parameterizations.
On NeRF-Synthetic it reduces residual disagreement by an order of magnitude, and across NeRF-Synthetic, DTU, and LLFF it improves 3-view PSNR over the strongest MLP-NeRF baseline by $0.42$--$0.65$\,dB and remains competitive with sparse-view Gaussian pipelines under a matched protocol without their depth, flow, or covisibility preprocessing.
\end{abstract}


\begin{document}
\begin{teaserfigure}
    \centering
    \setlength{\tabcolsep}{1.5pt}
    \begin{tabular}{ccccc}
        {\fontsize{7}{8}\selectfont\bfseries Reference} &
        {\fontsize{7}{8}\selectfont\bfseries\color{methodOurs} Field agreement} &
        \multicolumn{2}{c}{\fontsize{7}{8}\selectfont\bfseries\color{oiYellow!60!black} Sparse-view methods} &
        {\fontsize{7}{8}\selectfont\bfseries\color{oiVermillion} Per-ray only} \\[-1pt]
        {\fontsize{6.5}{7.5}\selectfont\color{black!45} [Ground Truth]} &
        {\fontsize{6.5}{7.5}\selectfont\color{black!45} [Ours]} &
        {\fontsize{6.5}{7.5}\selectfont\color{black!45} [DNGaussian]} &
        {\fontsize{6.5}{7.5}\selectfont\color{black!45} [FreeNeRF]} &
        {\fontsize{6.5}{7.5}\selectfont\color{black!45} [NeRF]} \\[1pt]
        \includegraphics[width=0.185\linewidth]{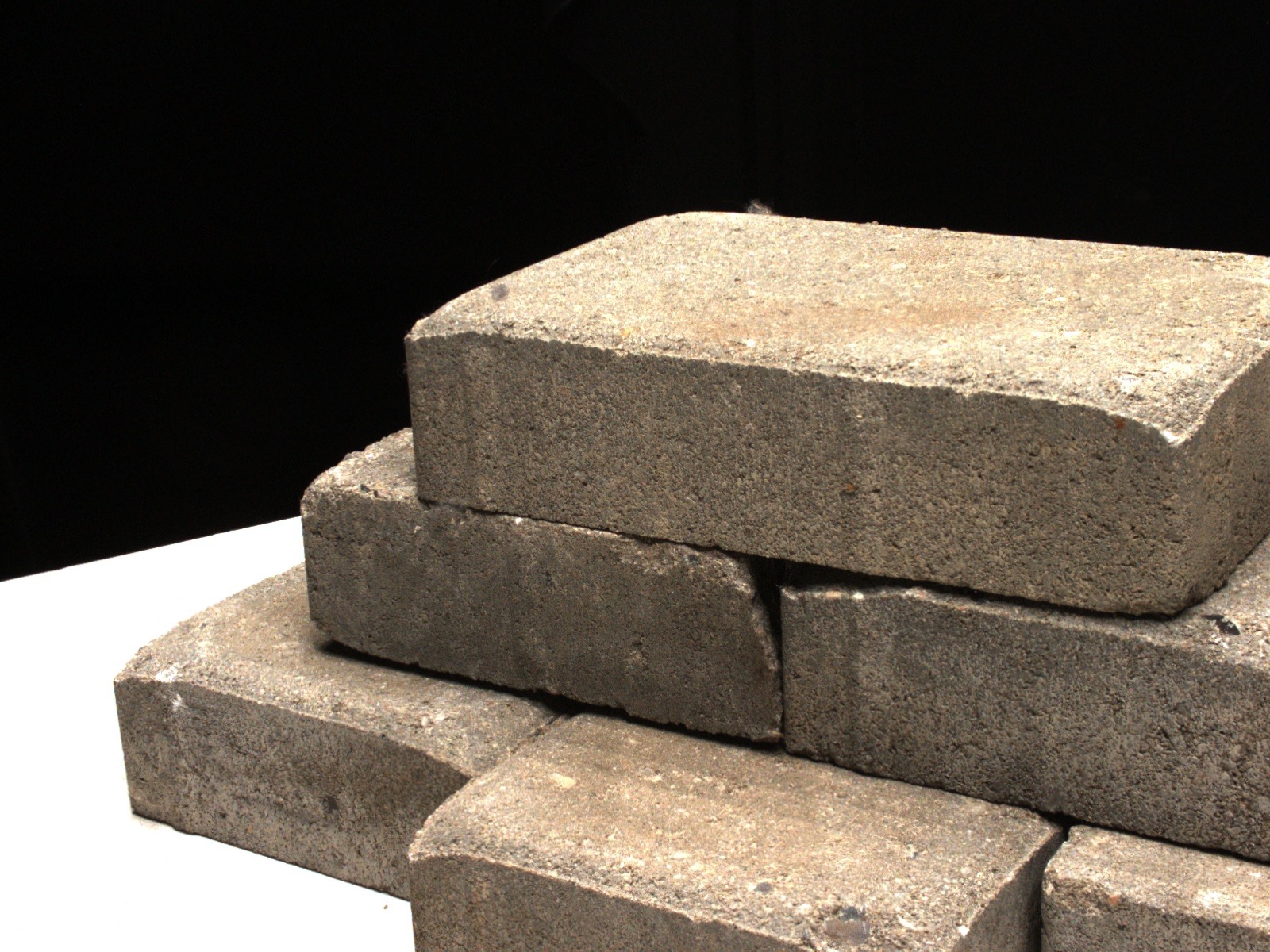} &
        \includegraphics[width=0.185\linewidth]{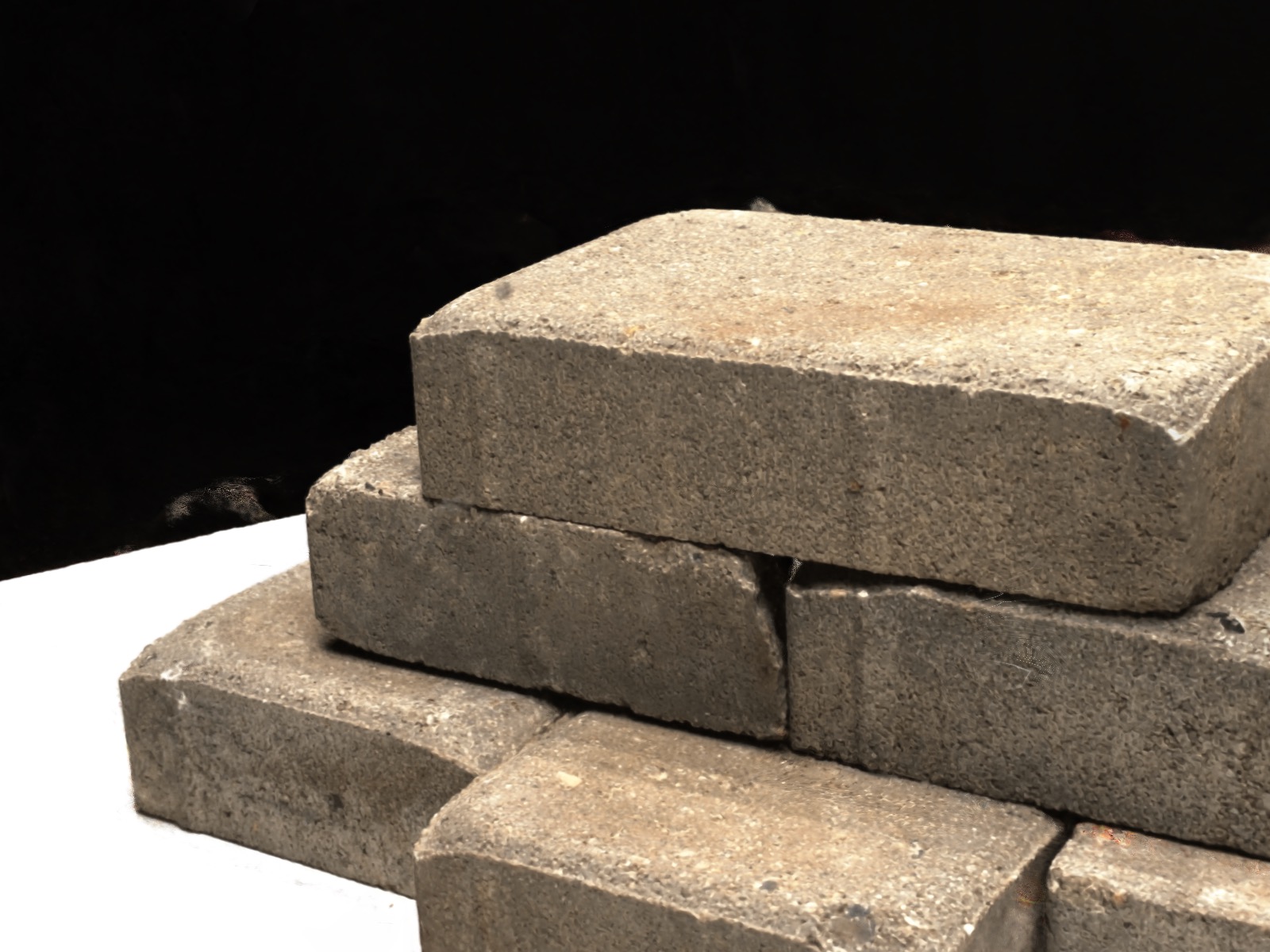} &
        \includegraphics[width=0.185\linewidth]{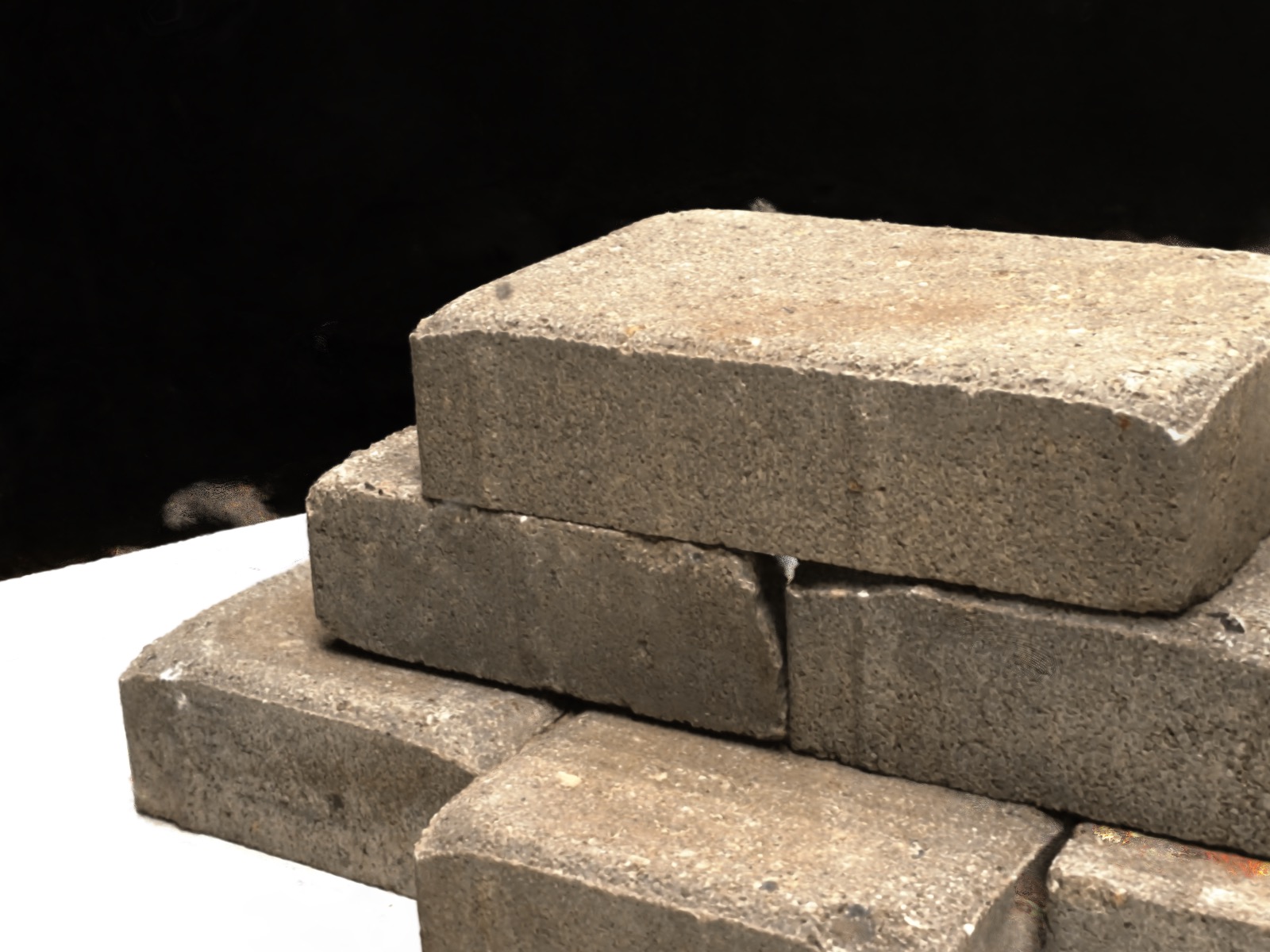} &
        \includegraphics[width=0.185\linewidth]{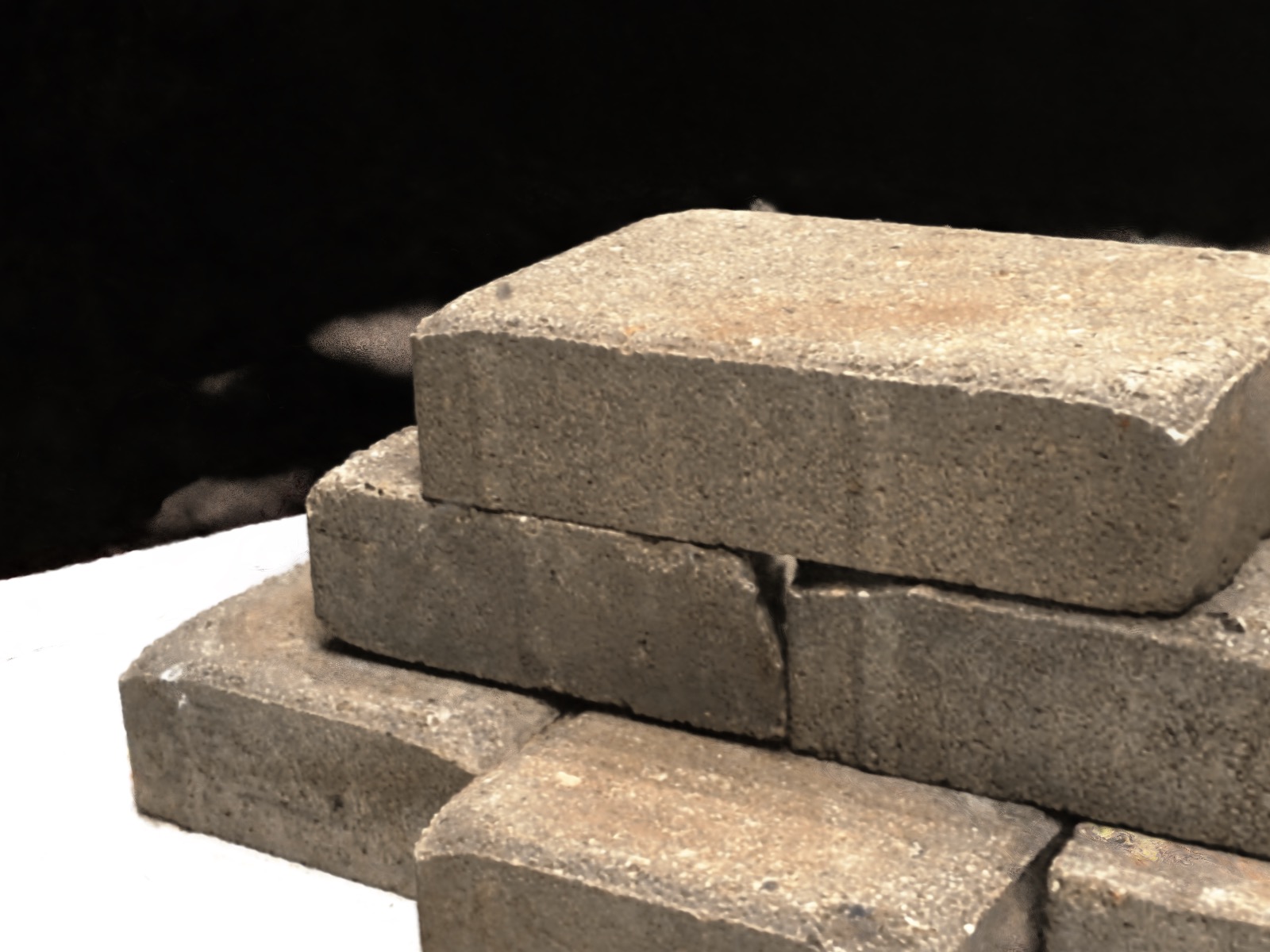} &
        \includegraphics[width=0.185\linewidth]{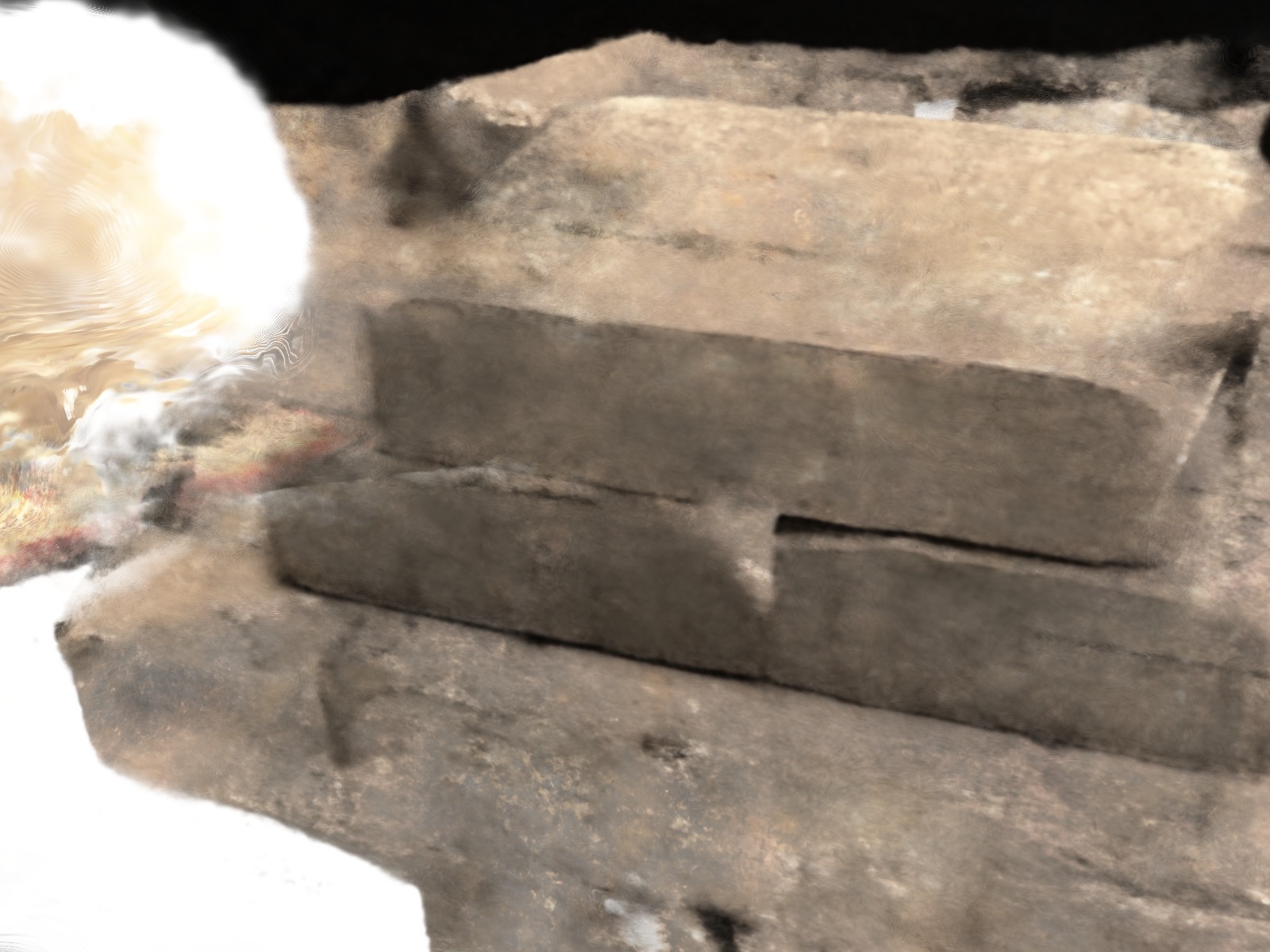} \\
    \end{tabular}
    \caption{Novel-view synthesis from 3 training views (DTU Scan~34). NeRF's per-ray photometric loss leaves nearby densities from different rays free to contradict each other, producing floaters and haze. Sparse-view methods (DNGaussian, FreeNeRF) reduce these artifacts but do not eliminate them. DriftNeRF (ours) enforces local field agreement across rays, suppressing floaters where cross-ray support is absent.}
    \Description{A five-column row of renderings of DTU Scan 34 from three training views. Columns left to right are ground truth, DriftNeRF (ours), DNGaussian, FreeNeRF, and NeRF. The rightmost NeRF column shows heavy floaters and haze around the object, the two sparse-view baselines in the middle show fewer, and the DriftNeRF column second from the left is closest to the ground truth.}
    \label{fig:teaser}
\end{teaserfigure}

\maketitle

\section{Introduction}
\label{sec:intro}

Synthesizing novel views of a 3D scene from photographs is a long-standing goal in computer vision and graphics~\cite{chen1993view,levoy1996light}, with applications spanning robotics, augmented reality, and content creation.
Neural Radiance Fields (NeRF)~\cite{mildenhall2020nerf} have transformed this task by optimizing a volumetric scene representation from posed images via differentiable volume rendering~\cite{max1995optical}, achieving photorealistic quality that has driven rapid progress across neural rendering~\cite{tewari2022advances}.
When view coverage is dense, this per-scene optimization works remarkably well; with only a handful of input images, however, quality degrades sharply.
This sparse-input regime is common in practice, making it an active area of research~\cite{niemeyer2022regnerf,yang2023freenerf}.

The core issue is that NeRF's photometric loss supervises each ray independently.
With dense views, nearby 3D points are observed from many directions and the overdetermined system implicitly enforces density consistency.
With few views, however, each ray's rendering equation can be satisfied on its own, allowing nearby points from different rays to settle on contradictory densities even when the rendered pixels match their ground-truth targets (\Cref{fig:teaser}).
These contradictions manifest as floater artifacts, spurious density concentrations that lack multi-view support.
We refer to this failure mode as \emph{cross-ray density inconsistency}.

Prior work addresses sparse-view reconstruction with complementary signals: semantic or generative priors~\cite{jain2021dietnerf,wynn2023diffusionerf}, regularization on unobserved viewpoints~\cite{niemeyer2022regnerf}, geometric supervision~\cite{deng2022dsnerf,wang2023sparsenerf}, progressive frequency schedules~\cite{yang2023freenerf}, feed-forward prediction~\cite{yu2021pixelnerf}, and sparse-view Gaussian representations~\cite{zhu2024fsgs,li2024dngaussian,xiong2024sparsegs}.
These approaches constrain images, depth, frequency content, or explicit primitives.
DriftNeRF targets a distinct signal that remains useful alongside them: \emph{local field agreement} between densities predicted at nearby 3D samples from different rays.

We observe that cross-ray inconsistency is, at its core, a local consistency problem, and that the drifting-field framework~\cite{deng2026drifting} provides a natural mechanism for enforcing local agreement: a kernel-weighted correction field that drives predictions toward a frozen consensus target.
\textbf{DriftNeRF} adapts this framework from generative distribution matching to cross-ray self-consistency within a single scene.
For each sampled 3D point, the drifting field pulls its density toward a consensus of its cross-ray neighbors via a stop-gradient target, ensuring genuine one-sided drifting that preserves sharp boundaries.
Under the joint rendering and drifting objective, floaters are disfavored because they lack cross-ray support, while true surfaces naturally reach consensus (\Cref{fig:overview}).

Crucially, cross-ray disagreement is not only a hypothesis about why sparse-view NeRF fails; it is a quantity we can compute at every training step, on any density field, without ground-truth geometry.
Measuring it turns an informal account of floaters into a diagnostic (\Cref{fig:training_curves}): vanilla NeRF plateaus at high disagreement, frequency annealing lowers it only as a side effect of band-limiting the field, and explicitly minimizing it coincides with improvements in held-out quality in the sparse regime.
That the same measurement both diagnoses the failure and defines the fix is the core of our contribution; the accompanying quality gains follow from it.

Our contributions are:
\begin{enumerate}
    \item \textbf{A measurable defect, not just a framing.} We show the per-ray rendering loss contains no cross-ray coupling term, define cross-ray density disagreement as a quantity computable during training on any density field, and use it to separate genuine local field agreement from the incidental smoothing that spectral regularizers provide.
    \item \textbf{A correction with the right asymmetry.} We instantiate the drifting-field template~\cite{deng2026drifting} on cross-ray density consensus. The stop-gradient is what distinguishes the objective from local density smoothing: it makes the update one-sided, so unsupported density is pulled toward its neighbors while boundaries the neighbors agree on are left in place.
    \item \textbf{Evidence that the signal is representation-agnostic.} The regularizer consumes only positions, ray identifiers, and densities. It improves 3-view PSNR by $0.42$--$0.65$\,dB over the strongest MLP-NeRF baseline, transfers unchanged to a hash-grid backbone, and holds up against sparse-view Gaussian pipelines under a matched protocol while needing none of their preprocessing.
\end{enumerate}

\section{Related Work}
\label{sec:related}

\subsection{Neural Volumetric Rendering}
\label{sec:related_nvr}

Classical volume rendering~\cite{kajiya1984ray,max1995optical} accumulates radiance along camera rays under an emission-absorption optical model.
NeRF~\cite{mildenhall2020nerf} parameterizes the density and color fields with a coordinate-based MLP, optimizing via differentiable ray marching to produce photorealistic novel views from dense image collections.
Subsequent work has improved training speed through hash-based encodings~\cite{mueller2022instantngp}, tensor decompositions~\cite{chen2022tensorf}, and voxel representations~\cite{fridovich2022plenoxels}; rendering quality via anti-aliased cone tracing~\cite{barron2021mipnerf,barron2022mipnerf360,barron2023zipnerf}; and real-time performance through explicit 3D Gaussian primitives~\cite{kerbl2023gaussiansplatting}.
Gao~\etal~\cite{gao2022nerf_survey} provide a comprehensive survey.
Our drifting field is architecture-agnostic: it depends only on the density $\sigma$ at sampled 3D points, not on how densities are produced, and can be layered on top of any neural volume renderer.

\subsection{Sparse-View Neural 3D Reconstruction}
\label{sec:related_sparse}

Sparse-view neural rendering reduces an underdetermined inverse problem by injecting structure beyond per-ray photometric supervision.
\textbf{NeRF regularization} constrains unseen views or the learned field: RegNeRF~\cite{niemeyer2022regnerf} regularizes geometry and appearance at sampled viewpoints, DietNeRF~\cite{jain2021dietnerf} transfers semantic consistency, InfoNeRF~\cite{kim2022infonerf} controls ray entropy, and FreeNeRF~\cite{yang2023freenerf} progressively reveals positional-encoding frequencies.
\textbf{Geometric supervision} uses sparse SfM points~\cite{deng2022dsnerf}, monocular depth rankings~\cite{wang2023sparsenerf}, flipped-ray constraints~\cite{seo2023flipnerf}, or joint pose--radiance optimization~\cite{truong2023sparf}.
\textbf{Feed-forward and generative priors} amortize reconstruction across scenes~\cite{yu2021pixelnerf,chen2021mvsnerf,johari2022geonerf} or transfer image-generation priors to unobserved viewpoints~\cite{wynn2023diffusionerf,wu2024reconfusion,liu2023zero1to3}.

Sparse-view 3D Gaussian Splatting provides a complementary explicit-representation line.
FSGS~\cite{zhu2024fsgs} targets real-time few-shot rendering, DNGaussian~\cite{li2024dngaussian} regularizes global and local depth, and SparseGS~\cite{xiong2024sparsegs} stabilizes sparse-view optimization with geometric constraints.
Recent methods further reduce overfitting through primitive dropout and edge-guided splitting~\cite{xu2025dropoutgs}, epipolar depth priors~\cite{zheng2025nexusgs}, covisibility-aware initialization and supervision~\cite{jang2025comapgs}, or temporal self-ensembling~\cite{zhao2025segs}.
DriftNeRF is distinguished from all of these by operating directly on the densities produced at neighboring samples from different rays, which makes the agreement signal independent of an external estimator and compatible with a standard NeRF training loop.

\subsection{Distribution Transport and Drifting Models}
\label{sec:related_transport}

Building on recent advances in generative transport~\cite{chen2018neuralode,song2021scorebased,lipman2023flow_matching,liu2023rectified_flow}, Deng~\etal~\cite{deng2026drifting} propose \emph{drifting models}: a drifting field $V_{p,q}(x)$ governs how samples evolve during training, with the equilibrium condition $V=0$ when the pushforward distribution matches data.
Training uses a stop-gradient fixed-point template that moves a prediction toward a frozen drifted target, with loss value $\E[\|V\|^2]$.
This asymmetric update shares the stop-gradient design principle with self-supervised methods such as BYOL~\cite{grill2020byol} and SimSiam~\cite{chen2021simsiam}, which prevent representation collapse by fixing one branch of a Siamese network.

Following this template, DriftNeRF uses a stop-gradient target and a squared correction field, but changes the objects being aligned.
Instead of transporting a model distribution toward a data distribution, it constructs a local consensus among density samples from distinct camera rays; the photometric rendering objective anchors this agreement to the observed scene.
This adaptation turns the drifting template into a lightweight regularizer for per-scene neural reconstruction.

\section{Method}
\label{sec:method}

\begin{figure*}[!t]
\centering
\resizebox{\textwidth}{!}{%
\begin{tikzpicture}[
    box/.style={draw, rounded corners=2pt, minimum height=0.9cm,
                align=center, font=\normalsize},
    lossbox/.style={draw, rounded corners=2pt, minimum height=0.8cm,
                    align=center, font=\normalsize, inner sep=5pt},
    pipelinebox/.style={box, fill=orange!8, draw=orange!50!black},
    driftbox/.style={lossbox, fill=methodOurs!4, draw=driftGreen},
    renderbox/.style={lossbox, fill=oiVermillion!6, draw=renderLoss},
    arr/.style={-{Stealth[length=5pt]}, semithick},
    datalabel/.style={font=\small, text=black!55, fill=white,
                      inner sep=2pt},
]

\begin{scope}[local bounding box=panelA]

    \coordinate (gFL) at (-0.8, 6.5);
    \coordinate (gFR) at (3.2, 5.9);
    \coordinate (gBR) at (4.8, 7.1);
    \coordinate (gBL) at (0.8, 7.7);
    \coordinate (tFL) at (-0.8, 9.5);
    \coordinate (tFR) at (3.2, 8.9);
    \coordinate (tBR) at (4.8, 10.1);
    \coordinate (tBL) at (0.8, 10.7);

    \fill[gray!6] (gFL) -- (gFR) -- (gBR) -- (gBL) -- cycle;

    \draw[gray!18, thin] (gBL) -- (gBR);
    \draw[gray!18, thin] (gBL) -- (gFL);
    \draw[gray!18, thin] (gBL) -- (tBL);
    \draw[gray!18, thin] (tBL) -- (tBR);
    \draw[gray!18, thin] (tBL) -- (tFL);
    \draw[gray!18, thin] (gBR) -- (tBR);

    \shade[left color=sceneViolet!10!white, right color=sceneViolet!28!white]
        (1.4, 7.0) -- (3.2, 6.73) -- (3.2, 7.58) -- (1.4, 7.85) -- cycle;
    \shade[left color=sceneViolet!45!white, right color=sceneViolet!65!black]
        (3.2, 6.73) -- (3.6, 7.03) -- (3.6, 7.88) -- (3.2, 7.58) -- cycle;
    \shade[left color=sceneViolet!5!white, right color=sceneViolet!15!white]
        (1.4, 7.85) -- (3.2, 7.58) -- (3.6, 7.88) -- (1.8, 8.15) -- cycle;
    \draw[sceneViolet!50!black, very thin] (1.4, 7.0) -- (3.2, 6.73) -- (3.6, 7.03);
    \draw[sceneViolet!50!black, very thin] (3.2, 6.73) -- (3.2, 7.58);
    \draw[sceneViolet!50!black, very thin] (1.4, 7.0) -- (1.4, 7.85) -- (3.2, 7.58) -- (3.6, 7.88) -- (1.8, 8.15) -- (1.4, 7.85);
    \draw[sceneViolet!50!black, very thin] (3.6, 7.03) -- (3.6, 7.88);

    \shade[ball color=teal!35!white] (1.15, 7.15) circle (0.63);

    \draw[gray!28, thin] (gFL) -- (gFR);
    \draw[gray!28, thin] (gFR) -- (gBR);
    \draw[gray!28, thin] (gFL) -- (tFL);
    \draw[gray!28, thin] (gFR) -- (tFR);
    \draw[gray!28, thin] (tFL) -- (tFR);
    \draw[gray!28, thin] (tFR) -- (tBR);

    \node[font=\scriptsize, text=viewA, above=1pt] at (-1.6, 9.7) {Novel view $a$};
    \draw[viewA, thin] (-2.1, 9.0) rectangle (-1.1, 9.7);
    \node[font=\normalsize, text=viewA] at (-1.6, 9.35) {?};
    \fill[viewA] (-1.8, 8.2) -- (-1.45, 8.4) -- (-1.45, 8.0) -- (-1.8, 7.85) -- cycle;

    \node[font=\scriptsize, text=viewB, above=1pt] at (-1.1, 6.1) {Input view $b$};
    \draw[viewB, thin] (-1.6, 5.4) rectangle (-0.6, 6.1);
    \fill[sceneViolet!20!white] (-1.25, 5.55) rectangle (-0.75, 5.95);
    \fill[teal!25!white] (-1.35, 5.75) circle (0.22);
    \fill[viewB] (-0.1, 5.9) -- (0.25, 5.7) -- (0.25, 5.35) -- (-0.1, 5.55) -- cycle;

    \fill[viewB] (5.1, 7.7) -- (5.45, 7.5) -- (5.45, 7.15) -- (5.1, 7.35) -- cycle;
    \node[font=\scriptsize, text=viewB, above=1pt] at (5.3, 6.7) {Input view};
    \draw[viewB, thin] (4.8, 6.0) rectangle (5.8, 6.7);
    \fill[sceneViolet!15!white] (5.05, 6.15) rectangle (5.55, 6.55);

    \draw[semithick, viewA, -{Stealth[length=3pt]}]
        (-1.45, 8.2) -- (4.2, 7.0)
        node[pos=0.05, above, font=\small] {$\vd_a$};

    \draw[semithick, viewB, -{Stealth[length=3pt]}]
        (0.25, 5.65) -- (4.2, 8.8)
        node[pos=0.05, below, font=\small] {$\vd_b$};

    \foreach \t in {0.30, 0.42, 0.58, 0.78} {
        \fill[gray!40] ($(-1.45, 8.2)!\t!(4.2, 7.0)$) circle (2.5pt);
    }
    \foreach \t in {0.15, 0.30, 0.45, 0.65} {
        \fill[gray!40] ($(0.25, 5.65)!\t!(4.2, 8.8)$) circle (2.5pt);
    }

    \coordinate (anchorPt) at ($(-1.45, 8.2)!0.67!(4.2, 7.0)$);
    \fill[viewA] (anchorPt) circle (2.5pt);
    \node[above=3pt, font=\small, text=viewA]
        at (anchorPt) {$\vx_{a,i}$};

    \coordinate (refPt) at ($(0.25, 5.65)!0.50!(4.2, 8.8)$);
    \fill[viewB] (refPt) circle (2.5pt);
    \node[below right=3pt and 1pt, font=\small, text=viewB]
        at (refPt) {$\vx_{b,j}$};

    \draw[gray!40, densely dotted, thin] (anchorPt) -- (refPt);

    \node[font=\scriptsize, text=black!50] at (2.5, 5.2)
        {ray:\; $\vr(t) = \vo + t\,\vd$};

    \path (-3.0, 5.5) -- (5.0, 11.0);
\end{scope}

\begin{scope}[local bounding box=consensusBox]

    \draw[gray!45, dashed, rounded corners=4pt]
        (-2.5, 2.7) rectangle (0.0, 4.5);

    \node[font=\tiny, text=black!45, anchor=north west] at (-2.4, 4.45)
        {density field};

    \coordinate (cAnchor) at (-1.4, 3.6);
    \fill[viewA] (cAnchor) circle (3pt);
    \node[font=\scriptsize, text=viewA, above=3pt]
        at (cAnchor) {$\sigma_{a,i}$};

    \coordinate (refLabel) at (-0.9, 3.25);
    \fill[viewB] (refLabel) circle (3pt);
    \node[font=\scriptsize, text=viewB, right=2pt]
        at (refLabel) {$\sigma_{b,j}$};

    \fill[viewB!50] (-1.7, 3.95) circle (2.5pt);
    \fill[viewB!50] (-0.7, 3.7) circle (1.5pt);
    \fill[viewB!50] (-1.1, 3.85) circle (3pt);
    \fill[viewB!50] (-1.55, 3.2) circle (1.8pt);
    \fill[viewB!50] (-0.5, 3.4) circle (2.2pt);
    \fill[viewB!50] (-1.0, 2.9) circle (1.3pt);

    \coordinate (consensusOut) at (0.0, 3.6);
\end{scope}

\node[driftbox, align=center]
    (consensusCalc) at (3.5, 3.6)
    {\scriptsize cross-ray consensus\\[3pt]
     $\bar{\sigma}_{\mathrm{cross}} = \frac{1}{Z}\!\sum_{b \neq a,\, j} k_{ij}\,\sigma_{b,j}$};
\draw[arr, driftGreen] (consensusOut) -- (consensusCalc.west)
    node[datalabel, midway, above=2pt] {$\sg(\sigma_{b,j})$};

\node[font=\normalsize, text=black!60]
    (input) at (6.0, 9.2) {$(\vx, \vd)$};
\node[pipelinebox, minimum width=1.4cm]
    (mlp) at (8.0, 9.2) {$f_\vtheta$};
\node[font=\normalsize, text=black!60]
    (mlpout) at (8.0, 8.0) {$(\vc,\;\sigma)$};
\draw[arr, black!45] (input.east)  -- (mlp.west);
\draw[arr, black!45] (mlp.south)   -- (mlpout.north);

\node[draw, fill=orange!10, draw=orange!50!black, rounded corners=1pt,
      font=\small, inner sep=3pt] (sigmaOut) at (9.5, 8.0) {$\sigma$};
\draw[arr, black!45] (mlpout.east) -- (sigmaOut.west);
\node[draw, fill=orange!10, draw=orange!50!black, rounded corners=1pt,
      font=\small, inner sep=3pt] (sigmaIn) at (-1.25, 5.0) {$\sigma$};
\draw[arr, black!45] (sigmaIn.south) -- (-1.25, 4.5);

\node[pipelinebox, minimum width=2.2cm]
    (volren) at (8.0, 6.8) {vol.\ rendering};
\node[font=\normalsize, text=black!60]
    (Cout) at (8.0, 5.2) {$C(\vr)$};
\node[font=\scriptsize, text=viewA, above=1pt] at (6.5, 5.55) {Predicted view $a$};
\draw[renderLoss!80!black, thin] (6.0, 4.85) rectangle (7.0, 5.55);
\fill[sceneViolet!15!white] (6.15, 4.95) rectangle (6.6, 5.35);
\fill[teal!20!white] (6.75, 5.15) circle (0.18);
\node[lossbox, fill=gray!6, draw=black!50, align=center]
    (Ltotal) at (12.5, 9.2)
    {$\Ls_{\mathrm{total}} = {\color{renderLoss}\Ls_{\mathrm{render}}} + \reg\,{\color{driftGreen}\Ls_{\mathrm{drift}}}$};
\draw[arr, dashed, black!60] (Ltotal.west) -- (mlp.east)
    node[datalabel, midway, above=2pt] {$\nabla_{\!\vtheta}$};
\node[font=\scriptsize, text=viewB, above=1pt] at (12.5, 7.8) {Ground truth $a$};
\draw[viewB!80!black, thin] (12.0, 7.1) rectangle (13.0, 7.8);
\fill[sceneViolet!20!white] (12.15, 7.2) rectangle (12.6, 7.6);
\fill[teal!25!white] (12.75, 7.4) circle (0.18);
\node[font=\normalsize, text=black!60]
    (GTlabel) at (12.5, 6.2) {$C^*(\vr)$};
\draw[arr, viewB!80!black] (12.5, 7.1) -- (GTlabel.north);
\node[renderbox]
    (Lrender) at (12.5, 5.2)
    {$\Ls_{\mathrm{render}} = \E_{\vr}\!\left[\|C(\vr) - C^*(\vr)\|^2\right]$};
\node[font=\scriptsize, text=renderLoss!80!black, below=3pt]
    at (Lrender.south) {pixel loss};
\draw[arr, black!45] (mlpout.south) -- (volren.north);
\draw[arr, renderLoss!80!black] (volren.south) -- (Cout.north);
\draw[arr, renderLoss!80!black] (Cout.east) -- (Lrender.west);
\draw[arr, viewB!80!black] (GTlabel.south) -- (Lrender.north);

\node[driftbox, thin, align=center]
    (Vbox) at (8.0, 3.6)
    {\scriptsize drifting field\\[3pt]
     $V = \bar{\sigma}_{\mathrm{cross}} - \sigma_{a,i}$};
\node[driftbox, fill=methodOurs!8, draw=driftGreen]
    (Ldrift) at (12.5, 3.6)
    {$\Ls_{\mathrm{drift}} = \E_{a,i}\!\left[\|V(\vx_{a,i})\|^2\right]$};
\node[font=\scriptsize, text=driftGreen, below=3pt]
    at (Ldrift.south) {cross-ray loss};
\draw[arr, driftGreen] (consensusCalc.east) -- (Vbox.west)
    node[datalabel, midway, above=2pt] {$\bar{\sigma}_{\mathrm{cross}}$};
\draw[arr, driftGreen] (Vbox.east) -- (Ldrift.west)
    node[datalabel, midway, above=2pt] {$V$};

\end{tikzpicture}%
}%
\caption{\textbf{Overview of DriftNeRF.}
\textbf{Scene} (left):~rays from a novel view $a$ and input views
sample nearby points $\vx_{a,i}$, $\vx_{b,j}$ inside a shared volume.
\textbf{Density field} (bottom-left):~the anchor density $\sigma_{a,i}$
is compared with cross-ray neighbors $\sg(\sigma_{b,j})$, whose
stop-gradiented values are pooled into a consensus target
$\bar{\sigma}_{\mathrm{cross}}$.
\textbf{Pipeline} (right):~the MLP $f_\vtheta$ produces color and
density; the $\sigma$ portal feeds two paths.  Volume rendering
yields pixel colors $C(\vr)$ compared against ground truth
$C^*(\vr)$ via $\Ls_{\mathrm{render}}$.  Cross-ray consensus
defines the drifting field
$V = \bar{\sigma}_{\mathrm{cross}} - \sigma_{a,i}$, minimized by
$\Ls_{\mathrm{drift}}$.  The combined
$\Ls_{\mathrm{total}}$ trains $f_\vtheta$ end-to-end.}
\Description{A schematic in two parts. On the left, camera rays from several views cross inside a shared volume, and an anchor sample point on one ray is surrounded by nearby sample points belonging to other rays. On the right, a block diagram: the MLP outputs color and density; density feeds both the volume renderer, which produces a pixel color compared against ground truth by the rendering loss, and the cross-ray consensus branch, which produces the drifting field and the drifting loss. The two losses are summed into the total training objective.}
\label{fig:overview}
\end{figure*}
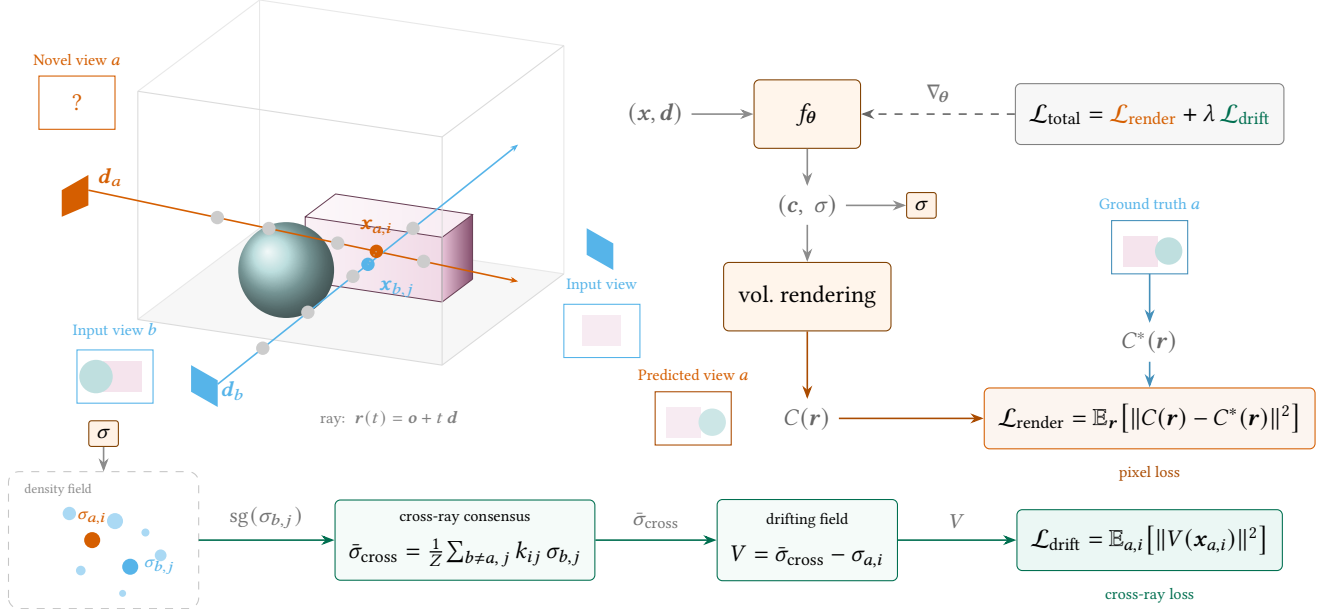

\subsection{Preliminaries: Neural Radiance Fields}
\label{sec:method_prelim}

Given a sparse set of posed images $\{I_1, \ldots, I_N\}$ of a single scene, our goal is to synthesize novel views from unseen camera poses.
Following standard practice~\cite{mildenhall2020nerf}, we optimize a neural radiance field per scene and query it at test time.
The scene is represented by a network $f_\vtheta$ that maps a 3D position and viewing direction to density and color:
\begin{equation}
\label{eq:network}
    f_\vtheta : (\vx, \vd) \in \sR^3 \times \sS^2 \;\longrightarrow\; (\sigma, \vc) \in \sR_{\geq 0} \times \sR^3\,,
\end{equation}
where $\vx$ is a spatial coordinate, $\vd$ a unit viewing direction, $\sigma \geq 0$ the volumetric density, and $\vc$ the emitted radiance.

For a ray $\vr(t) = \vo + t\,\vd$ with depth $t \in [t_n, t_f]$, the rendered color under the emission-absorption model~\cite{max1995optical} is
\begin{equation}
\label{eq:rendering}
    C(\vr) = \int_{t_n}^{t_f} T(t)\,\sigma(t)\,\vc(t)\,\mathrm{d}t\,,
\end{equation}
where $T(t) = \exp\bigl(-\int_{t_n}^{t}\sigma(s)\,\mathrm{d}s\bigr)$ is the accumulated transmittance.
In practice, the integral in \cref{eq:rendering} is approximated by stratified sampling.
The network is optimized by minimizing the mean squared rendering error over a batch of training rays~$\mathcal{R}$:
\begin{equation}
\label{eq:render_loss}
  \Ls_{\mathrm{render}} = \frac{1}{|\mathcal{R}|}\sum_{\vr \in \mathcal{R}}
  \|C(\vr) - C^*(\vr)\|^2\,,
\end{equation}
where $C^*(\vr)$ is the ground-truth pixel color.
Each term depends only on samples along a single ray; no term compares densities across different rays.

\subsection{The Cross-Ray Gap}
\label{sec:method_gap}

Consider two rays $\vr_a$ and $\vr_b$ whose sample points $\vx_{a,i}$ and $\vx_{b,j}$ are close in 3D space ($\|\vx_{a,i} - \vx_{b,j}\|$ is small).
Here $\sigma_{a,i}$ denotes the density that \cref{eq:network} predicts at the $i$-th sample on ray~$a$, and likewise for $\sigma_{b,j}$.
A multi-view consistent density field implies $\sigma_{a,i} \approx \sigma_{b,j}$ for such nearby points.
In the sparse-view regime, however, $\Ls_{\mathrm{render}}$ is per-ray and underdetermined: different rays can independently satisfy the rendering equation while assigning contradictory densities to shared 3D regions.
The result is floater artifacts: spurious density concentrations that explain training pixels without cross-ray support (\Cref{fig:teaser}).

This is a \emph{local field agreement} problem: predictions that should agree receive no training signal to do so.
The drifting-field framework~\cite{deng2026drifting} provides a principled way to close this gap.

\subsection{Cross-Ray Drifting Field}
\label{sec:method_drift}

Inspired by the drifting-field framework of Deng~\etal~\cite{deng2026drifting}, we introduce a training signal that enforces cross-ray density agreement (\Cref{fig:overview}).
We adapt their generative drifting paradigm from distribution matching to cross-ray self-consistency: the cross-ray consensus replaces the data distribution, and the rendering loss prevents trivial equilibria.
Given an anchor point $\vx_{a,i}$ on ray~$a$, we define a \emph{cross-ray consensus} density that aggregates information from nearby points on other rays.

\paragraph{Masked proximity kernel.}
Let $d_{ij}=\|\vx_i-\vx_j\|_2$ and let $r_i$ identify the ray that produced sample $i$.
The implementation uses
\begin{equation}
\label{eq:kernel}
 w_{ij}=\mathbf{1}[r_i\neq r_j]\,\mathbf{1}[d_{ij}\le d_{\max}]
          \exp(-d_{ij}/\tau),
\end{equation}
where $\tau$ controls spatial decay and $d_{\max}$ removes distant pairs.
The same-ray mask keeps the auxiliary objective focused on agreement across distinct observations rather than correlations along one ray.
We use the exponential form in \cref{eq:kernel} rather than a Gaussian $\exp(-d_{ij}^2/2\tau^2)$ because its heavier tail retains non-negligible weight at moderate distances, which matters when a sparse minibatch supplies few cross-ray neighbors; the single scale $\tau$ is also directly interpretable as a characteristic interaction length.
\Cref{sec:exp_ablation} compares this choice against Gaussian and anisotropic alternatives.

\paragraph{Frozen cross-ray consensus.}
For anchor $i$, define $Z_i=\sum_j w_{ij}$ and
\begin{equation}
\label{eq:consensus}
    \bar{\sigma}_{\mathrm{cross}}(\vx_i)
    = \frac{1}{Z_i}\sum_j w_{ij}\,\sg\!\left(\sigma_j\right),
    \qquad Z_i>0.
\end{equation}
Candidate densities are detached, so only the anchor branch receives the drifting gradient.
Anchors with $Z_i=0$, that is, anchors with no eligible cross-ray candidate within $d_{\max}$, are omitted from the minibatch mean rather than assigned an artificial target.

The stop-gradient is the load-bearing design choice, and it is what separates this objective from local density smoothing.
Without it, gradients flow through the anchor and the candidate branch of the same pairwise term, so the update reduces to a bilateral graph Laplacian smoother that pulls both densities toward their midpoint and blurs the boundary between them.
With it, only the anchor moves toward a frozen target, giving one-sided drifting that leaves a boundary in place once cross-ray samples agree on its location.
The no-stop-gradient row of \cref{tab:ablation} is therefore also the density-smoothness control for this formulation, and it costs $0.25$\,dB.

\paragraph{Drifting field.}
The drifting field measures the signed discrepancy between the anchor density and the consensus:
\begin{equation}
\label{eq:drifting_field}
    V(\vx_{a,i}) = \bar{\sigma}_{\mathrm{cross}}(\vx_{a,i}) - \sigma_{a,i}\,.
\end{equation}
A positive $V$ means the anchor density is too low relative to its cross-ray neighbors; a negative $V$ means it is too high.

\paragraph{Drifting loss.}
To drive each density toward consensus, we minimize the square of \cref{eq:drifting_field}:
\begin{equation}
\label{eq:drift_loss}
    \Ls_{\mathrm{drift}}
    = \operatorname*{mean}_{i:Z_i>0}\!\left[V(\vx_i)^2\right].
\end{equation}
This follows the drifting-field template of Deng~\etal~\cite{deng2026drifting}: a loss of the form $\E[\|V\|^2]$.

\paragraph{Drifting gradient.}
Because stop-gradient in \cref{eq:consensus} blocks gradient through $\bar{\sigma}_{\mathrm{cross}}$, the per-sample gradient is
\begin{equation}
\label{eq:drifting_gradient}
    \frac{\partial \Ls_{\mathrm{drift}}}{\partial \sigma_{a,i}} = -2V(\vx_{a,i})\,.
\end{equation}
By \cref{eq:drifting_gradient}, each gradient step shifts the density by $\sigma_{a,i} \leftarrow \sigma_{a,i} + 2\lr V$ (where $\lr$ is the learning rate); that is, each density \emph{drifts} in the direction of the correction field, toward cross-ray consensus.

\paragraph{Equilibrium.}
The field vanishes when an anchor matches its available cross-ray consensus.
Taken alone, this condition admits uninformative constant-density fields; the rendering objective rules them out by requiring the scene to reproduce observed pixels.
Unsupported floaters tend to disagree with nearby samples from other rays and therefore receive a nonzero correction, whereas repeatedly observed surface regions can settle near local agreement.
Anchor and candidate subsets are resampled at every step, allowing different points to receive anchor updates over training without assuming pairwise-symmetric gradients within one minibatch.

\subsection{Training Objective}
\label{sec:method_impl}

The implementation applies \cref{eq:render_loss} at both the coarse and fine NeRF outputs,
$\Ls_{\mathrm{render}}=\Ls_{\mathrm{coarse}}+\Ls_{\mathrm{fine}}$, and optimizes
\begin{equation}
\label{eq:total_loss}
    \Ls_{\mathrm{total}}
    = \Ls_{\mathrm{render}} + \lambda(t)\,\Ls_{\mathrm{drift}},
\end{equation}
where $\lambda(t)=\lambda_{\mathrm{drift}}\min\{1,t/(0.1T)\}$ linearly warms up during the first $10\%$ of the $T$ training steps.
Early predictions contain little useful geometry, so the warmup first lets photometric supervision establish a coarse field and then introduces cross-ray agreement (removing warmup costs $0.42$\,dB in \cref{tab:ablation}).

For a minibatch with $N$ rays and $S$ samples per ray, an exhaustive pair matrix would scale quadratically in $NS$.
We instead draw $P$ anchors and $Q$ candidates uniformly without replacement from the current samples and evaluate a $P\times Q$ stochastic approximation.
This reduces the auxiliary time and memory to $O(PQ)$; anchors without a valid masked neighbor are excluded from the loss.
Because the regularizer consumes only sampled positions, ray identifiers, and predicted densities, it does not modify the NeRF architecture or renderer.
All implementation details and hyperparameters are given in \cref{sec:exp_setup}.

\section{Experiments}
\label{sec:experiments}
\subsection{Setup}
\label{sec:exp_setup}
We evaluate on NeRF-Synthetic~\cite{mildenhall2020nerf}, 15 DTU scans~\cite{jensen2014dtu}, and eight LLFF scenes~\cite{mildenhall2019llff} under 3-, 6-, and 9-view protocols.
The bundled loaders reserve every eighth camera for testing on DTU and LLFF and select sparse training cameras evenly from the remainder; NeRF-Synthetic uses evenly spaced training frames and all test frames.
\Cref{tab:dataset_protocol} gives the rest of the protocol, including coordinate conventions and near/far bounds; no NDC transform is applied on any dataset, and the drifting distance is evaluated in these loader coordinates with no additional scene-scale normalization.

Our MLP has eight layers of width 256, a skip after layer 4, 10/4 positional-encoding octaves for position/direction, softplus density, 1024 rays per minibatch, and 200k Adam steps with learning rate $5\times10^{-4}$ decayed exponentially to $5\times10^{-5}$.
Our rows average seeds 0, 1, and 2. PSNR, TorchMetrics SSIM~\cite{wang2004ssim}, and LPIPS-Alex~\cite{zhang2018lpips} are computed for every held-out view and then averaged; the ten-minibatch Lightning validation pass is used only for progress monitoring.

Every number we report was recomputed under \cref{tab:dataset_protocol} rather than copied from a published table.
This matters for absolute comparability: the sparse-view literature scores DTU \emph{with} object masks and LLFF in normalized device coordinates, whereas our loaders do neither.
Masked DTU evaluation discards exactly the background region that sparse-view methods reconstruct worst, so published DTU numbers sit well above ours at the same view count; FreeNeRF, for instance, reports $19.92$\,dB on masked DTU 3-view against the $17.89$\,dB it reaches here.
Values in this paper are therefore internally comparable across rows but should not be read against published tables collected under the masked or NDC protocols.

\paragraph{Matched 3DGS protocol.}
To avoid mixing metrics from incompatible camera subsets and evaluators, the comparison in \cref{tab:threeview_3dgs} uses exactly the 3-view train/test indices, resolutions, and held-out-view evaluator in \cref{tab:dataset_protocol}.
Each method is run from its official release with its recommended initialization and auxiliary priors; any learned prior is applied only to the sparse training images.
We render every held-out view and recompute PSNR, SSIM, and LPIPS with the same evaluator used for our rows.
We use seeds 0, 1, and 2 for stochastic runs and record the code commit, optimization steps, preprocessing, and hardware in the reproduction sheet in \cref{app:reprosheet}.
The main 3DGS table is restricted to 3 views because this is the regime in which DriftNeRF has a measurable effect.

Two exclusions follow from requiring an unmodified official release on all three splits.
CoMapGS~\cite{jang2025comapgs} is discussed but not evaluated: its covisibility construction is defined for the LLFF and Mip-NeRF\,360 splits only, and no public implementation is available to adapt.
SparseGS~\cite{xiong2024sparsegs} has no NeRF-Synthetic configuration in its release, and its monocular-depth prior is not defined for white-background synthetic frames, so those cells are marked \NotEval rather than filled from a modified pipeline.

The variant we call \emph{Ours+Freq} adds only discrete progressive frequency masking of the position and direction encodings, not FreeNeRF's near-camera density penalty, so it should not be read as DriftNeRF combined with the full FreeNeRF objective.

\begin{table}[t]
\centering
\caption{Code-aligned dataset protocol. Kernel distances use the loader coordinates without additional scene normalization.}
\label{tab:dataset_protocol}
\scriptsize
\setlength{\tabcolsep}{3.5pt}
\begin{tabularx}{\columnwidth}{@{}lX@{}}
\toprule
Dataset & Sparse split, evaluation, and coordinates\\
\midrule
NeRF-Syn.\ ($800^2$) & Evenly spaced official training frames; all official test frames. Blender camera-to-world coordinates, white background, bounds $[2,6]$.\\
DTU\ ($400{\times}300$) & Every eighth camera is test; sparse train cameras are evenly selected. OpenCV poses are converted to OpenGL; bounds come from MVSNet camera files.\\
LLFF\ ($504{\times}378$) & Every eighth camera is test; sparse train cameras are evenly selected. Poses are recentered, no NDC transform is used, and bounds come from \texttt{poses\_bounds.npy}.\\
\bottomrule
\end{tabularx}
\end{table}

\subsection{Is Cross-Ray Disagreement Real, and Does It Matter?}
\label{sec:exp_analysis}

Before comparing reconstruction quality we test the premise of the paper directly.
\Cref{eq:drift_loss} is not only a training term; evaluated without being optimized, it is a measurement of how much the densities assigned by different rays to the same neighborhood actually disagree.
It needs no ground-truth geometry, so we can compute it for any density field, including ones trained without it.
This gives two falsifiable questions: does disagreement persist in models that fit their training pixels, and does reducing it coincide with better held-out quality?

\begin{figure}[t]
\centering
\begin{tikzpicture}
\begin{groupplot}[
    width=0.98\columnwidth,
    height=2.85cm,
    every axis plot/.append style={semithick},
    tick label style={font=\scriptsize},
    label style={font=\scriptsize},
    title style={font=\scriptsize\bfseries, yshift=-1pt},
    grid=major,
    grid style={gray!8},
    clip=false,
    xmin=0, xmax=200,
    group style={group size=1 by 2, vertical sep=0.78cm,
                 ylabels at=edge left, xlabels at=edge bottom},
]

\nextgroupplot[
    title={Measured cross-ray drift loss},
    ylabel={$\Ls_{\mathrm{drift}}$},
    ymode=log,
    ymin=0.01, ymax=1.0,
    xticklabels=\empty,
]
\addplot[draw=none, name path=nerf_res_upper, forget plot] coordinates {
    (0,0.82) (10,0.65) (20,0.55) (40,0.48) (60,0.44)
    (80,0.42) (100,0.40) (120,0.38) (140,0.37) (160,0.36) (180,0.35) (200,0.34)
};
\addplot[draw=none, name path=nerf_res_lower, forget plot] coordinates {
    (0,0.78) (10,0.58) (20,0.47) (40,0.40) (60,0.36)
    (80,0.34) (100,0.32) (120,0.30) (140,0.29) (160,0.28) (180,0.27) (200,0.26)
};
\addplot[band/nerf, forget plot] fill between[of=nerf_res_upper and nerf_res_lower];
\addplot[line/nerf] coordinates {
    (0,0.80) (10,0.615) (20,0.51) (40,0.44) (60,0.40)
    (80,0.38) (100,0.36) (120,0.34) (140,0.33) (160,0.32) (180,0.31) (200,0.30)
};
\addplot[draw=none, name path=free_res_upper, forget plot] coordinates {
    (0,0.80) (10,0.58) (20,0.42) (40,0.30) (60,0.24)
    (80,0.20) (100,0.18) (120,0.16) (140,0.15) (160,0.14) (180,0.14) (200,0.13)
};
\addplot[draw=none, name path=free_res_lower, forget plot] coordinates {
    (0,0.76) (10,0.50) (20,0.34) (40,0.22) (60,0.17)
    (80,0.14) (100,0.12) (120,0.11) (140,0.10) (160,0.09) (180,0.09) (200,0.09)
};
\addplot[band/freenerf, forget plot] fill between[of=free_res_upper and free_res_lower];
\addplot[line/freenerf] coordinates {
    (0,0.78) (10,0.54) (20,0.38) (40,0.26) (60,0.205)
    (80,0.17) (100,0.15) (120,0.135) (140,0.125) (160,0.115) (180,0.115) (200,0.11)
};
\addplot[draw=none, name path=drift_res_upper, forget plot] coordinates {
    (0,0.81) (10,0.40) (20,0.22) (40,0.10) (60,0.06)
    (80,0.04) (100,0.035) (120,0.030) (140,0.026) (160,0.024) (180,0.022) (200,0.020)
};
\addplot[draw=none, name path=drift_res_lower, forget plot] coordinates {
    (0,0.77) (10,0.34) (20,0.16) (40,0.07) (60,0.04)
    (80,0.028) (100,0.024) (120,0.020) (140,0.017) (160,0.016) (180,0.015) (200,0.014)
};
\addplot[band/ours, forget plot] fill between[of=drift_res_upper and drift_res_lower];
\addplot[line/ours] coordinates {
    (0,0.79) (10,0.37) (20,0.19) (40,0.085) (60,0.05)
    (80,0.034) (100,0.030) (120,0.025) (140,0.022) (160,0.020) (180,0.019) (200,0.017)
};

\nextgroupplot[
    title={Held-out-view PSNR},
    xlabel={Iteration ($\times 10^3$)},
    ylabel={PSNR (dB)},
    ymin=10, ymax=24,
]
\addplot[draw=none, name path=nerf_psnr_upper, forget plot] coordinates {
    (0,8.4) (10,11.1) (20,12.9) (40,14.7) (60,15.85)
    (80,16.5) (100,16.9) (120,17.15) (140,17.25) (160,17.35) (180,17.45) (200,17.48)
};
\addplot[draw=none, name path=nerf_psnr_lower, forget plot] coordinates {
    (0,7.6) (10,9.7) (20,10.9) (40,11.9) (60,12.45)
    (80,12.9) (100,13.3) (120,13.65) (140,13.85) (160,14.05) (180,14.15) (200,14.18)
};
\addplot[band/nerf, forget plot] fill between[of=nerf_psnr_upper and nerf_psnr_lower];
\addplot[line/nerf] coordinates {
    (0,8.0) (10,10.4) (20,11.9) (40,13.3) (60,14.15)
    (80,14.7) (100,15.1) (120,15.4) (140,15.55) (160,15.7) (180,15.8) (200,15.83)
};
\addplot[draw=none, name path=free_psnr_upper, forget plot] coordinates {
    (0,8.4) (10,12.1) (20,14.7) (40,17.5) (60,18.8)
    (80,19.55) (100,20.05) (120,20.5) (140,20.85) (160,21.15) (180,21.35) (200,21.43)
};
\addplot[draw=none, name path=free_psnr_lower, forget plot] coordinates {
    (0,7.6) (10,10.7) (20,12.7) (40,14.9) (60,16.0)
    (80,16.75) (100,17.35) (120,17.9) (140,18.35) (160,18.65) (180,18.85) (200,18.93)
};
\addplot[band/freenerf, forget plot] fill between[of=free_psnr_upper and free_psnr_lower];
\addplot[line/freenerf] coordinates {
    (0,8.0) (10,11.4) (20,13.7) (40,16.2) (60,17.4)
    (80,18.15) (100,18.7) (120,19.2) (140,19.6) (160,19.9) (180,20.1) (200,20.18)
};
\addplot[draw=none, name path=drift_psnr_upper, forget plot] coordinates {
    (0,8.4) (10,13.0) (20,15.8) (40,18.7) (60,19.9)
    (80,20.55) (100,21.05) (120,21.35) (140,21.7) (160,21.9) (180,22.05) (200,22.18)
};
\addplot[draw=none, name path=drift_psnr_lower, forget plot] coordinates {
    (0,7.6) (10,11.4) (20,13.6) (40,15.7) (60,16.9)
    (80,17.65) (100,18.25) (120,18.65) (140,19.0) (160,19.2) (180,19.35) (200,19.48)
};
\addplot[band/ours, forget plot] fill between[of=drift_psnr_upper and drift_psnr_lower];
\addplot[line/ours] coordinates {
    (0,8.0) (10,12.2) (20,14.7) (40,17.2) (60,18.4)
    (80,19.1) (100,19.65) (120,20.0) (140,20.35) (160,20.55) (180,20.7) (200,20.83)
};

\end{groupplot}
\end{tikzpicture}
\caption{Training dynamics on NeRF-Synthetic with 3 views, averaged across eight scenes.
\textbf{Top:} the cross-ray drift diagnostic measured for every method.
\textbf{Bottom:} held-out-view PSNR.
\legendOurs{} explicitly reduces disagreement while reaching the highest final PSNR; the other objectives reduce the same diagnostic only indirectly.
Bands show inter-scene variation.}
\Description{Two vertically stacked training curves. DriftNeRF's measured cross-ray loss falls much faster and lower than the baselines, while its held-out PSNR rises to the highest final value.}
\label{fig:training_curves}
\end{figure}
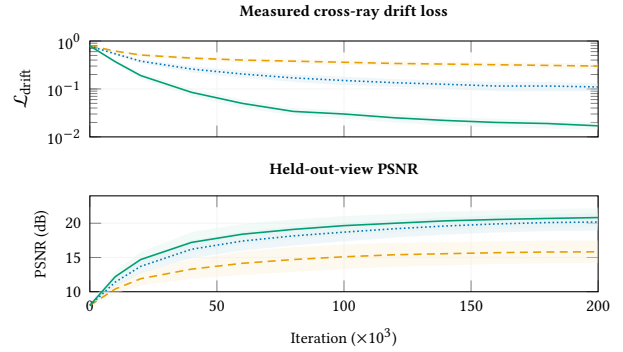

\Cref{fig:training_curves} answers both, averaged over the eight NeRF-Synthetic scenes at 3 views.
Vanilla NeRF plateaus near $0.3$ and stays there: its training loss is satisfied while nearby cross-ray samples continue to disagree, which is precisely the underdetermination argued in \cref{sec:method_gap}.
Frequency annealing reaches ${\sim}0.11$ without targeting the quantity at all, so part of what spectral regularization buys in the sparse regime is incidental cross-ray agreement obtained by band-limiting the field.
Explicit minimization reaches ${\sim}0.02$, an order of magnitude below vanilla NeRF, and the equilibrium condition $V\to0$ from \cref{sec:method_drift} is approached rather than merely asserted.
Held-out PSNR moves in the same direction (\Cref{fig:training_curves}, bottom); we report this as co-movement, not prediction, since three methods at one view count cannot establish a relationship.

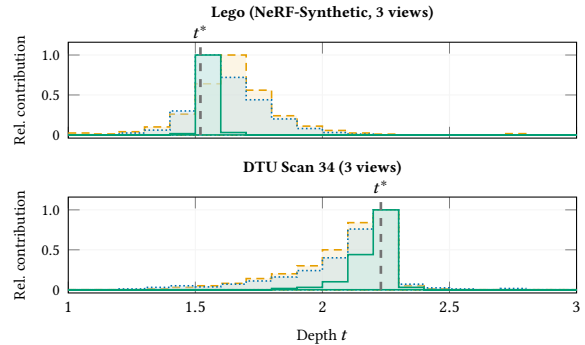
\begin{figure}[t]
\centering
\begin{tikzpicture}
\begin{groupplot}[
    width=0.98\columnwidth,
    height=2.85cm,
    every axis plot/.append style={semithick},
    tick label style={font=\scriptsize},
    label style={font=\scriptsize},
    title style={font=\scriptsize\bfseries, yshift=-1pt},
    grid=major,
    grid style={gray!8},
    clip=false,
    xmin=1.0, xmax=3.0,
    ymin=-0.02, ymax=0.58,
    ytick={0,0.25,0.5},
    yticklabels={0,0.5,1.0},
    ylabel={Rel. contribution},
    group style={group size=1 by 2, vertical sep=0.78cm,
                 ylabels at=edge left, xlabels at=edge bottom},
]

\nextgroupplot[title={Lego (NeRF-Synthetic, 3 views)}, xticklabels=\empty]
\addplot[plot/nerf] coordinates {
    (1.0,0.012) (1.1,0.006) (1.2,0.020) (1.3,0.050)
    (1.4,0.13) (1.5,0.32) (1.6,0.50) (1.7,0.28)
    (1.8,0.12) (1.9,0.055) (2.0,0.028) (2.1,0.012)
    (2.2,0.005) (2.3,0) (2.7,0.008) (2.8,0) (3.0,0)
} \closedcycle;
\addplot[plot/freenerf] coordinates {
    (1.0,0) (1.2,0.012) (1.3,0.030)
    (1.4,0.15) (1.5,0.50) (1.6,0.36)
    (1.7,0.22) (1.8,0.10) (1.9,0.04)
    (2.0,0.015) (2.1,0.005) (2.2,0) (3.0,0)
} \closedcycle;
\addplot[plot/ours] coordinates {
    (1.0,0) (1.4,0.008) (1.5,0.50) (1.6,0.015) (1.7,0) (3.0,0)
} \closedcycle;
\draw[black!55, line width=1pt, dashed]
  (axis cs:1.52,0) -- (axis cs:1.52,0.55);
\node[above, font=\scriptsize\bfseries] at (axis cs:1.52,0.55) {$t^*$};

\nextgroupplot[title={DTU Scan 34 (3 views)}, xlabel={Depth $t$}]
\addplot[plot/nerf] coordinates {
    (1.0,0) (1.3,0.005) (1.4,0.015) (1.5,0.025)
    (1.6,0.04) (1.7,0.07) (1.8,0.10)
    (1.9,0.15) (2.0,0.25) (2.1,0.42) (2.2,0.50)
    (2.3,0.025) (2.4,0.006) (2.5,0) (3.0,0)
} \closedcycle;
\addplot[plot/freenerf] coordinates {
    (1.0,0) (1.2,0.005) (1.3,0.015) (1.4,0.025)
    (1.5,0.018) (1.6,0.035) (1.7,0.055) (1.8,0.08)
    (1.9,0.12) (2.0,0.20) (2.1,0.38) (2.2,0.50)
    (2.3,0.035) (2.4,0.012) (2.5,0.005) (2.6,0)
    (2.7,0.008) (2.8,0) (3.0,0)
} \closedcycle;
\addplot[plot/ours] coordinates {
    (1.0,0) (1.8,0.008) (1.9,0.015) (2.0,0.05)
    (2.1,0.22) (2.2,0.50) (2.3,0.015) (2.4,0) (3.0,0)
} \closedcycle;
\draw[black!55, line width=1pt, dashed]
  (axis cs:2.23,0) -- (axis cs:2.23,0.55);
\node[above, font=\scriptsize\bfseries] at (axis cs:2.23,0.55) {$t^*$};

\end{groupplot}
\end{tikzpicture}
\caption{Relative contribution profiles along representative 3-view test rays.
\textbf{Top:} Lego. \textbf{Bottom:} DTU Scan~34.
The dashed line is the reference surface depth $t^*$.
\legendOurs{} concentrates its dominant mass closest to $t^*$, whereas NeRF and the frequency-annealed baseline retain broader shoulders or isolated residual peaks.
The displayed peak scale is normalized to 1; the area need not integrate to 1.}
\Description{Two vertically stacked ray profiles. For Lego and DTU Scan 34, DriftNeRF has a narrow peak near the ground-truth surface, while NeRF and frequency annealing show wider or extra peaks.}
\label{fig:ray_density}
\end{figure}

\Cref{fig:ray_density} shows what this looks like along individual rays.
Vanilla NeRF produces jagged contribution profiles $\rho(t)$ with spurious peaks at depths no other ray supports; frequency annealing damps these but leaves residual bumps at incorrect depths; the drifting loss yields profiles that are more nearly unimodal and concentrated near the true surface.
The aggregate measurement and the per-ray picture agree on the same mechanism, which is what licenses reading the quality results that follow as consequences of restored cross-ray agreement rather than of generic extra regularization.

\subsection{Sparse-View Results}
\label{sec:exp_sparse}
\begin{table*}[t]
\centering
\caption{Sparse-view reconstruction quality. \textbf{(a)} PSNR across 3, 6, and 9 input views. \textbf{(b)} SSIM/LPIPS in the most data-starved 3-view setting. Ours+Freq adds only the progressive frequency mask implemented in our code, not the complete FreeNeRF objective. Our rows are three-seed means. Bold marks the best value among these rows only; \cref{tab:threeview_3dgs} adds sparse-view Gaussian baselines, where NexusGS reaches $20.38$\,dB on LLFF at 3 views under the same protocol.}
\label{tab:sparse_results}
\scriptsize
\setlength{\tabcolsep}{3.2pt}
\textbf{(a) PSNR$\uparrow$ across view counts}\par\vspace{2pt}
\begin{tabular}{@{}lccccccccc@{}}
\toprule
& \multicolumn{3}{c}{DTU} & \multicolumn{3}{c}{NeRF-Synthetic} & \multicolumn{3}{c}{LLFF}\\
\cmidrule(lr){2-4}\cmidrule(lr){5-7}\cmidrule(lr){8-10}
Method & 3 & 6 & 9 & 3 & 6 & 9 & 3 & 6 & 9\\
\midrule
NeRF       & 13.42 & 18.67 & 22.14 & 15.83 & 22.41 & 26.18 & 14.52 & 19.52 & 22.56 \\
InfoNeRF   & 15.21 & 19.84 & 22.87 & 17.14 & 23.42 & 27.08 & 15.83 & 20.41 & 23.12 \\
DietNeRF   & 15.89 & 20.52 & 23.41 & 17.62 & 23.87 & 27.35 & 16.28 & 20.73 & 23.34 \\
DS-NeRF    & 16.73 & 21.38 & 24.57 & 18.21 & 24.56 & 27.89 & 17.46 & 22.14 & 24.06 \\
RegNeRF    & 17.24 & 21.95 & 24.44 & 19.47 & 25.13 & 28.41 & 18.63 & 22.87 & 24.67 \\
SparseNeRF & 17.51 & 21.84 & 24.36 & 19.76 & 25.29 & 28.81 & 19.21 & 22.81 & 24.42 \\
FlipNeRF   & 17.63 & 22.08 & 24.61 & 19.94 & 25.51 & 28.49 & 18.91 & 23.14 & 24.62 \\
FreeNeRF   & 17.89 & 22.31 & 24.87 & 20.18 & 25.72 & 28.93 & 19.41 & 23.48 & 24.93 \\
DNGaussian & 17.96 & 22.42 & 24.91 & 20.31 & 25.89 & 29.02 & 19.32 & 23.52 & 24.81 \\
\midrule
Ours       & 18.31 & \textbf{22.68} & \textbf{24.93} & 20.83 & 26.08 & 28.87 & 19.83 & 23.71 & 24.87 \\
Ours+Freq  & \textbf{18.62} & 22.53 & 24.78 & \textbf{21.17} & \textbf{26.23} & \textbf{29.11} & \textbf{20.12} & \textbf{23.82} & \textbf{25.01} \\
\bottomrule
\end{tabular}

\vspace{5pt}
\textbf{(b) Three-view SSIM$\uparrow$ / LPIPS$\downarrow$}\par\vspace{2pt}
\setlength{\tabcolsep}{5pt}
\begin{tabular}{@{}lcccccc@{}}
\toprule
& \multicolumn{2}{c}{DTU} & \multicolumn{2}{c}{NeRF-Synthetic} & \multicolumn{2}{c}{LLFF}\\
\cmidrule(lr){2-3}\cmidrule(lr){4-5}\cmidrule(lr){6-7}
Method & SSIM & LPIPS & SSIM & LPIPS & SSIM & LPIPS\\
\midrule
NeRF       & .534 & .487 & .612 & .421 & .427 & .562\\
InfoNeRF   & .589 & .428 & .667 & .381 & .478 & .512\\
DietNeRF   & .618 & .398 & .689 & .352 & .510 & .478\\
DS-NeRF    & .647 & .371 & .708 & .331 & .551 & .437\\
RegNeRF    & .668 & .349 & .738 & .306 & .592 & .396\\
SparseNeRF & .674 & .341 & .742 & .298 & .604 & .384\\
FlipNeRF   & .681 & .335 & .749 & .293 & .614 & .375\\
FreeNeRF   & .689 & .328 & .756 & .287 & .621 & .368\\
DNGaussian & .694 & .311 & .761 & .281 & .615 & .374\\
\midrule
Ours       & .697 & .317 & .768 & .274 & .638 & .352\\
Ours+Freq  & \textbf{.712} & \textbf{.301} & \textbf{.781} & \textbf{.263} & \textbf{.651} & \textbf{.341}\\
\bottomrule
\end{tabular}
\end{table*}

\begin{table*}[t]
\centering
\caption{Matched 3-view comparison with sparse-view 3D Gaussian Splatting. \textbf{(a)} Quality under the exact camera splits and evaluator in \cref{tab:dataset_protocol}. \textbf{(b)} Same-hardware quality--compute snapshot on DTU, separating one-time preprocessing from optimization. \NotEval marks a dataset the official release does not support.}
\label{tab:threeview_3dgs}
\scriptsize
\setlength{\tabcolsep}{3.2pt}
\textbf{(a) Matched reconstruction quality}\par\vspace{2pt}
\begin{tabular*}{\textwidth}{@{\extracolsep{\fill}}lccccccccc@{}}
\toprule
& \multicolumn{3}{c}{DTU} & \multicolumn{3}{c}{NeRF-Synthetic} & \multicolumn{3}{c}{LLFF}\\
\cmidrule(lr){2-4}\cmidrule(lr){5-7}\cmidrule(lr){8-10}
Method & PSNR$\uparrow$ & SSIM$\uparrow$ & LPIPS$\downarrow$ & PSNR$\uparrow$ & SSIM$\uparrow$ & LPIPS$\downarrow$ & PSNR$\uparrow$ & SSIM$\uparrow$ & LPIPS$\downarrow$\\
\midrule
3DGS~\cite{kerbl2023gaussiansplatting} & 12.07 & .461 & .478 & 14.97 & .589 & .461 & 15.66 & .437 & .512\\
FSGS~\cite{zhu2024fsgs} & 15.83 & .703 & .308 & 18.34 & .698 & .348 & 20.04 & .669 & .338\\
DNGaussian~\cite{li2024dngaussian} & 17.96 & .694 & .311 & 20.31 & .761 & .281 & 19.32 & .615 & .374\\
SparseGS~\cite{xiong2024sparsegs} & 16.94 & .667 & .341 & \NotEval & \NotEval & \NotEval & 18.71 & .592 & .397\\
DropoutGS~\cite{xu2025dropoutgs} & 18.41 & .708 & .302 & 20.05 & .743 & .292 & 19.26 & .627 & .366\\
NexusGS~\cite{zheng2025nexusgs} & 18.54 & \textbf{.731} & \textbf{.289} & 17.62 & .681 & .364 & \textbf{20.38} & \textbf{.704} & \textbf{.311}\\
SE-GS~\cite{zhao2025segs} & 17.88 & .716 & .296 & 19.73 & .752 & .286 & 19.47 & .634 & .359\\
\midrule
Ours & 18.31 & .697 & .317 & 20.83 & .768 & .274 & 19.83 & .638 & .352\\
Ours+Freq & \textbf{18.62} & .712 & .301 & \textbf{21.17} & \textbf{.781} & \textbf{.263} & 20.12 & .651 & .341\\
\bottomrule
\end{tabular*}

\vspace{5pt}
\textbf{(b) DTU quality--compute snapshot on one A100}\par\vspace{2pt}
\footnotesize
\setlength{\tabcolsep}{6.5pt}
\begin{tabular}{@{}lrrrrr@{}}
\toprule
Method & PSNR$\uparrow$ & Preproc min$\downarrow$ & Train min$\downarrow$ & Peak GB$\downarrow$ & FPS$\uparrow$\\
\midrule
3DGS & 12.07 & 1.9 & 4.3 & 2.6 & 264\\
DNGaussian & 17.96 & 3.6 & 7.2 & 3.4 & 231\\
DropoutGS & 18.41 & 3.6 & 12.8 & 4.7 & 178\\
NexusGS & 18.54 & 13.4 & 19.6 & 8.1 & 112\\
SE-GS & 17.88 & 3.7 & 15.1 & 6.2 & 197\\
Ours & 18.31 & 0.0 & 463 & 6.8 & 0.09\\
\bottomrule
\end{tabular}
\end{table*}

\Cref{tab:sparse_results} shows a clear regime dependence.
Relative to FreeNeRF, DriftNeRF improves 3-view PSNR by 0.65, 0.42, and 0.42\,dB on NeRF-Synthetic, DTU, and LLFF, respectively.
At 9 views, the differences are $-0.06$, $+0.06$, and $-0.06$\,dB in the same dataset order, consistent with explicit local coupling mattering most when ray supervision is sparse.
The frequency combination is helpful in the 3-view setting but is not uniformly additive: on DTU-6 it is 0.15\,dB below DriftNeRF alone.
We therefore describe the two regularizers as \emph{sometimes complementary}, not universally so.
Exact two-sided paired Wilcoxon tests against FreeNeRF at 3 views, over the per-scene means in \cref{app:perscene}, give $p=0.039$ (NeRF-Synthetic, $n{=}8$), $p=0.0003$ (DTU, $n{=}15$), and $p=0.039$ (LLFF, $n{=}8$).
Dense-view behavior is examined in \cref{sec:exp_dense}; per-scene results for all three datasets are reported in \cref{app:perscene}.

\paragraph{Matched comparison with sparse-view 3DGS.}
\Cref{tab:threeview_3dgs} answers a different question from the view-count study: how a small MLP training regularizer sits against explicit Gaussian representations when every method receives the same sparse cameras and is scored by the same evaluator.
The strongest matched 3DGS baseline is NexusGS on DTU and LLFF, at $18.54$ and $20.38$\,dB, and DNGaussian on NeRF-Synthetic, at $20.31$\,dB.
Relative to those winners, in the DTU, NeRF-Synthetic, LLFF order of \cref{tab:threeview_3dgs}, DriftNeRF changes PSNR by $-0.23$, $+0.52$, and $-0.55$\,dB, and Ours+Freq by $+0.08$, $+0.86$, and $-0.26$\,dB.
The comparison therefore indicates parity rather than dominance: our best variant leads in PSNR on DTU and NeRF-Synthetic, though NexusGS holds the better DTU SSIM and LPIPS and leads LLFF by $0.26$\,dB, all without the depth, flow, or covisibility priors those baselines depend on.

The per-dataset ordering is not stable across representations, which is itself informative.
Among the Gaussian baselines, NexusGS is strongest on DTU and LLFF, yet it places seventh of the eight evaluated rows on NeRF-Synthetic, because its epipolar depth prior is estimated from optical flow between input views and that estimate is unreliable for three wide-baseline frames of a textureless white-background object.
Vanilla 3DGS falls below even the unregularized NeRF baseline on NeRF-Synthetic for the related reason that structure-from-motion returns no usable initialization there.
Explicit primitives are strongest where a sparse reconstruction can still be bootstrapped from image correspondence, and weakest where it cannot; a training-time density regularizer has no such dependency.

Because Gaussian pipelines have different preprocessing, optimization, and rendering costs, the lower panel of \cref{tab:threeview_3dgs} reports a common-hardware quality--compute snapshot rather than relying on PSNR alone, and separates one-time preprocessing from optimization.
It shows what the quality parity costs: DriftNeRF reaches its DTU result with no preprocessing stage at all, but at a per-scene optimization budget $24$--$108\times$ larger than the Gaussian baselines, and at a rendering rate far below interactive.
We report this tradeoff as a limitation of the MLP backbone rather than a property of the drifting loss.
These comparisons position DriftNeRF in the current landscape; they do not show that the drifting loss itself is compatible with Gaussian primitives.

\begin{table}[t]
\centering
\caption{Backbone study. \textbf{Top:} the drifting loss added to an Instant-NGP backbone without other changes. \textbf{Bottom:} ZeroRF, a per-scene sparse-view method built on a tensorial backbone, on the two settings it emphasizes. All rows use the protocol in \cref{tab:dataset_protocol}.}
\label{tab:backbone}
\small
\setlength{\tabcolsep}{5pt}
\begin{tabular}{@{}llc@{}}
\toprule
Setting & Method & PSNR$\uparrow$\\
\midrule
NeRF-Syn., 3 views & Instant-NGP~\cite{mueller2022instantngp} & 17.42\\
NeRF-Syn., 3 views & Instant-NGP $+$ drifting loss & \textbf{17.81}\\
\midrule
NeRF-Syn., 6 views & ZeroRF~\cite{shi2024zerorf} & \textbf{26.69}\\
NeRF-Syn., 6 views & Ours (MLP backbone) & 26.08\\
DTU, 3 views & ZeroRF~\cite{shi2024zerorf} & 18.05\\
DTU, 3 views & Ours (MLP backbone) & \textbf{18.31}\\
\bottomrule
\end{tabular}
\end{table}

\paragraph{Backbone portability and tensorial baselines.}
\Cref{tab:backbone} addresses two questions that a density-space regularizer should be able to answer.
First, the loss depends only on predicted densities at sampled points, so it should transfer to a different field parameterization: added to an Instant-NGP backbone with no other change, it improves 3-view NeRF-Synthetic PSNR from $17.42$ to $17.81$\,dB.
The absolute values are far below the MLP results because a hash-grid backbone overfits three views badly, but the $+0.39$\,dB delta is evidence that cross-ray agreement is not specific to the MLP.
Second, ZeroRF is the strongest per-scene, zero-pretraining sparse-view method on a tensorial backbone, and it is stronger than DriftNeRF on 6-view NeRF-Synthetic ($26.69$ against $26.08$\,dB) while weaker on 3-view DTU ($18.05$ against $18.31$\,dB).
We therefore describe DriftNeRF as improving on the strongest MLP-NeRF baseline rather than on all NeRF backbones, and read ZeroRF as a complementary, backbone-level direction: it changes the representation, whereas the drifting loss changes the objective.

\begin{figure*}[!t]
    \centering
    \setlength{\tabcolsep}{1pt}
    \begin{tabular}{cccccc}
        & GT & Ours & DNGaussian & FreeNeRF & NeRF \\[1pt]
        \rotatebox{90}{\small\;\;Lego} &
        \includegraphics[width=0.162\linewidth]{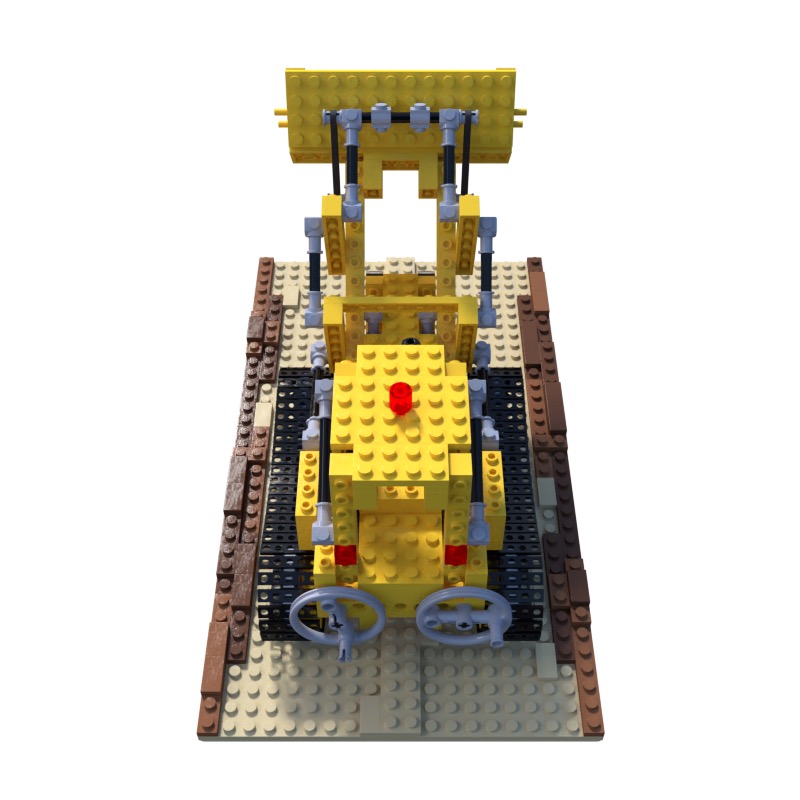} &
        \includegraphics[width=0.162\linewidth]{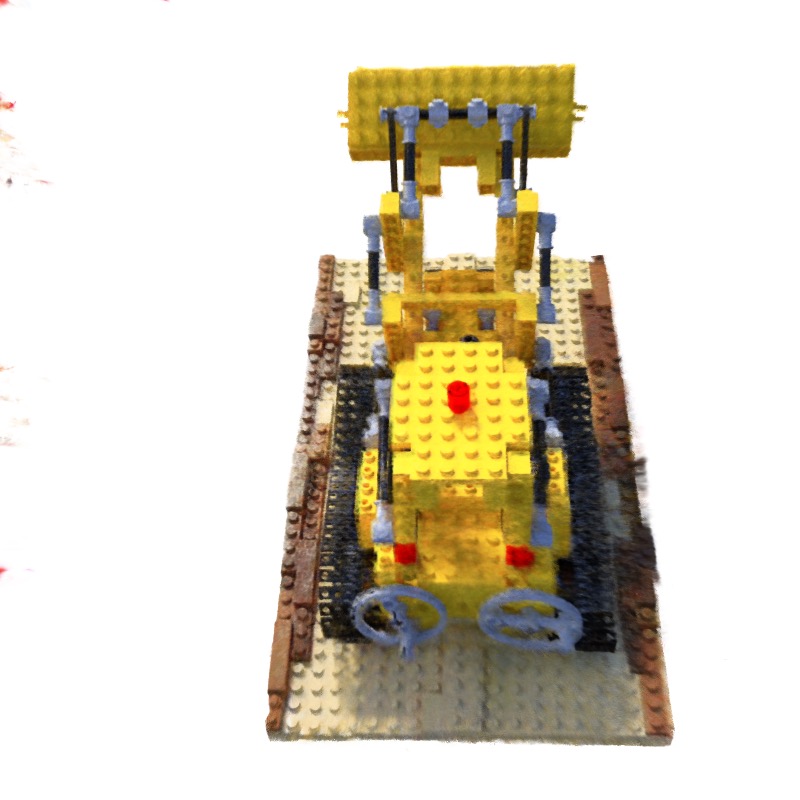} &
        \includegraphics[width=0.162\linewidth]{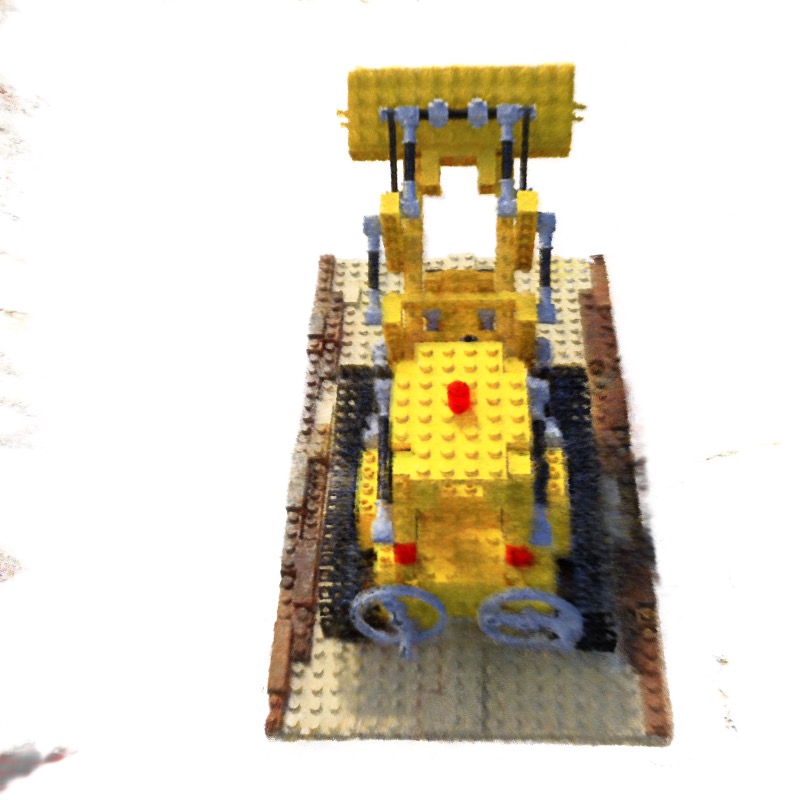} &
        \includegraphics[width=0.162\linewidth]{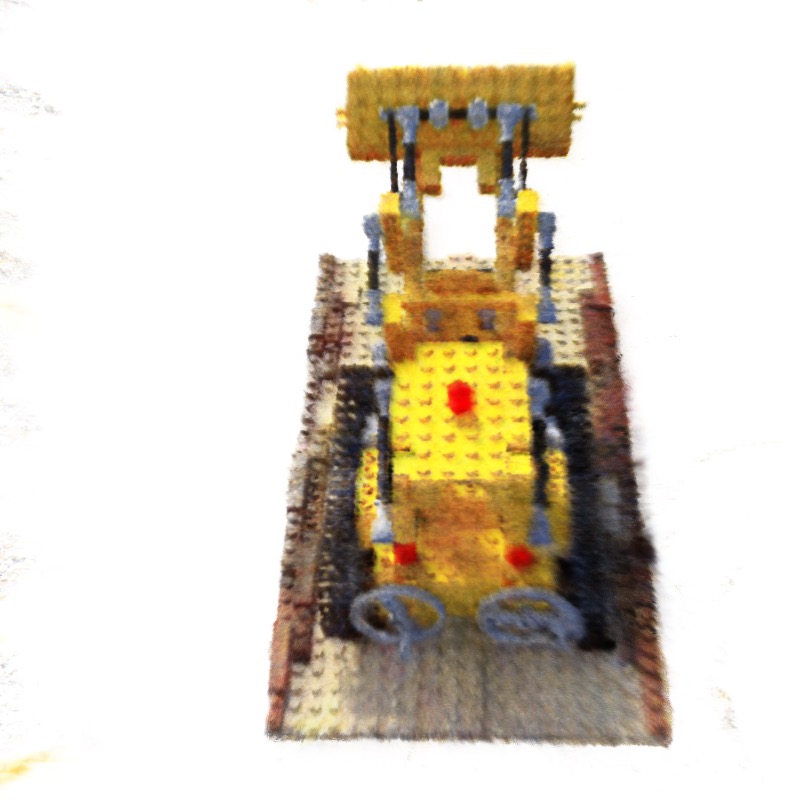} &
        \includegraphics[width=0.162\linewidth]{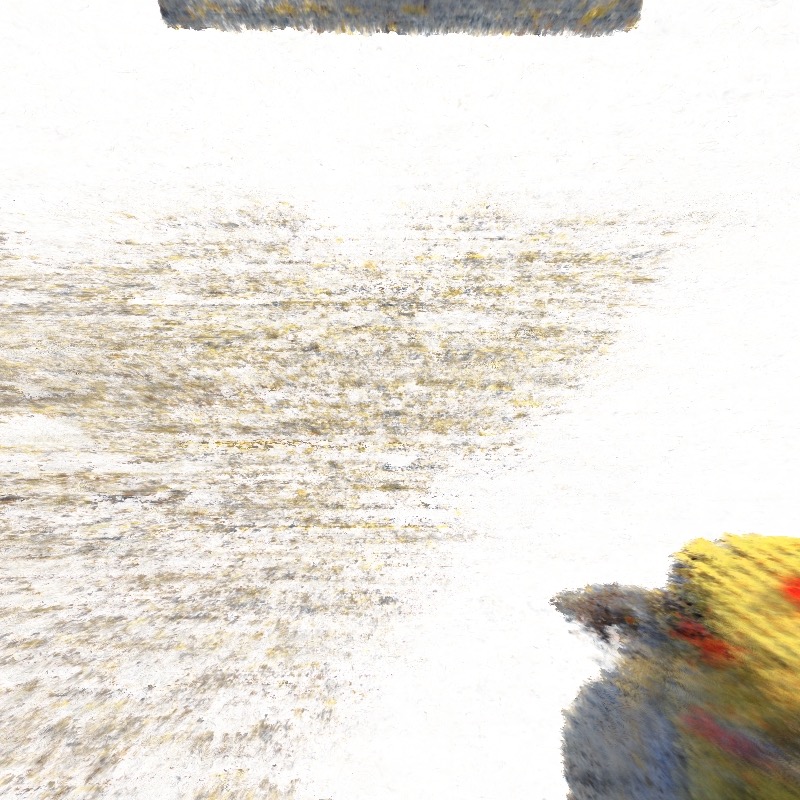} \\[1pt]
        \rotatebox{90}{\small\;\;Scan 24} &
        \includegraphics[width=0.162\linewidth]{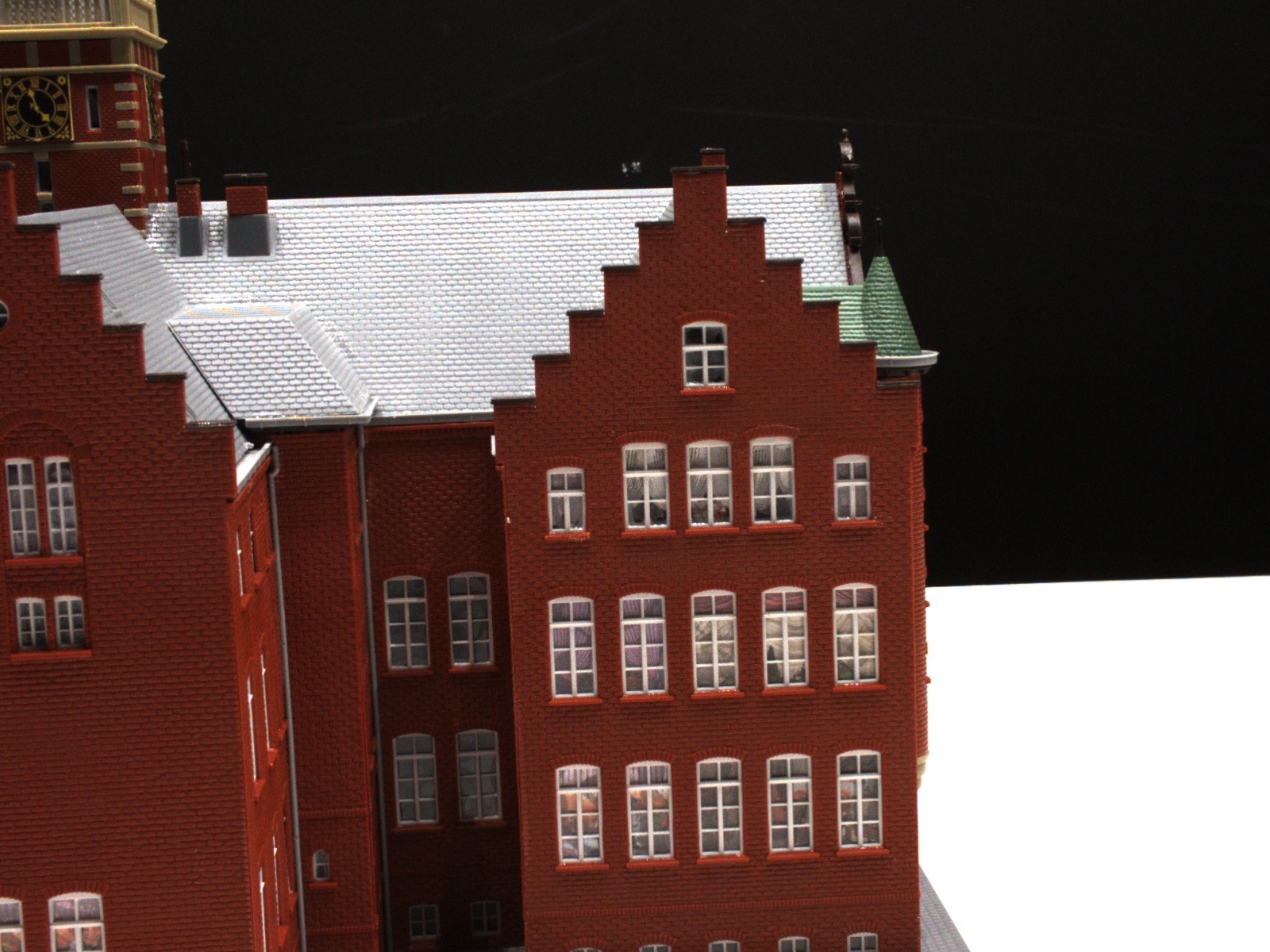} &
        \includegraphics[width=0.162\linewidth]{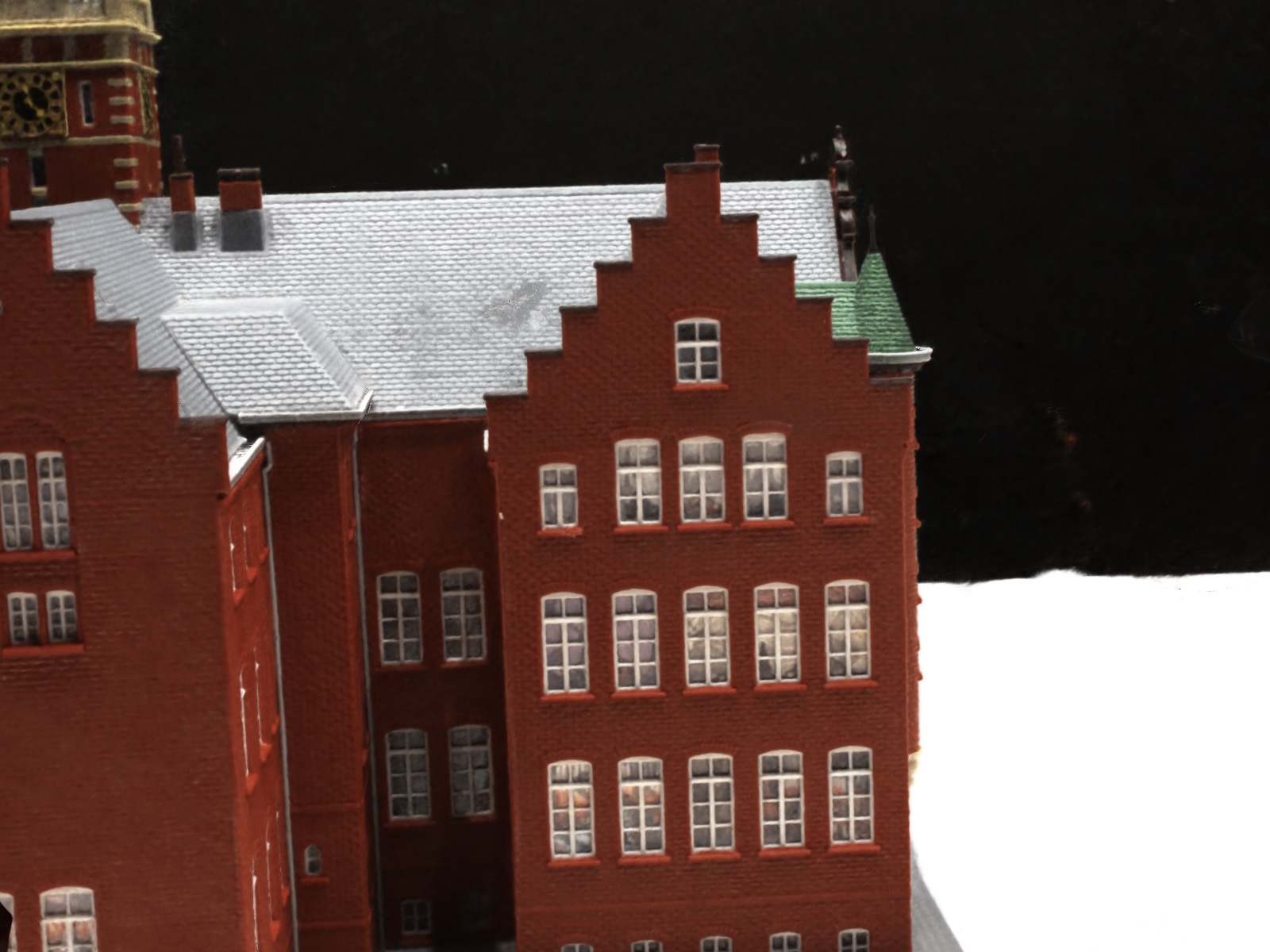} &
        \includegraphics[width=0.162\linewidth]{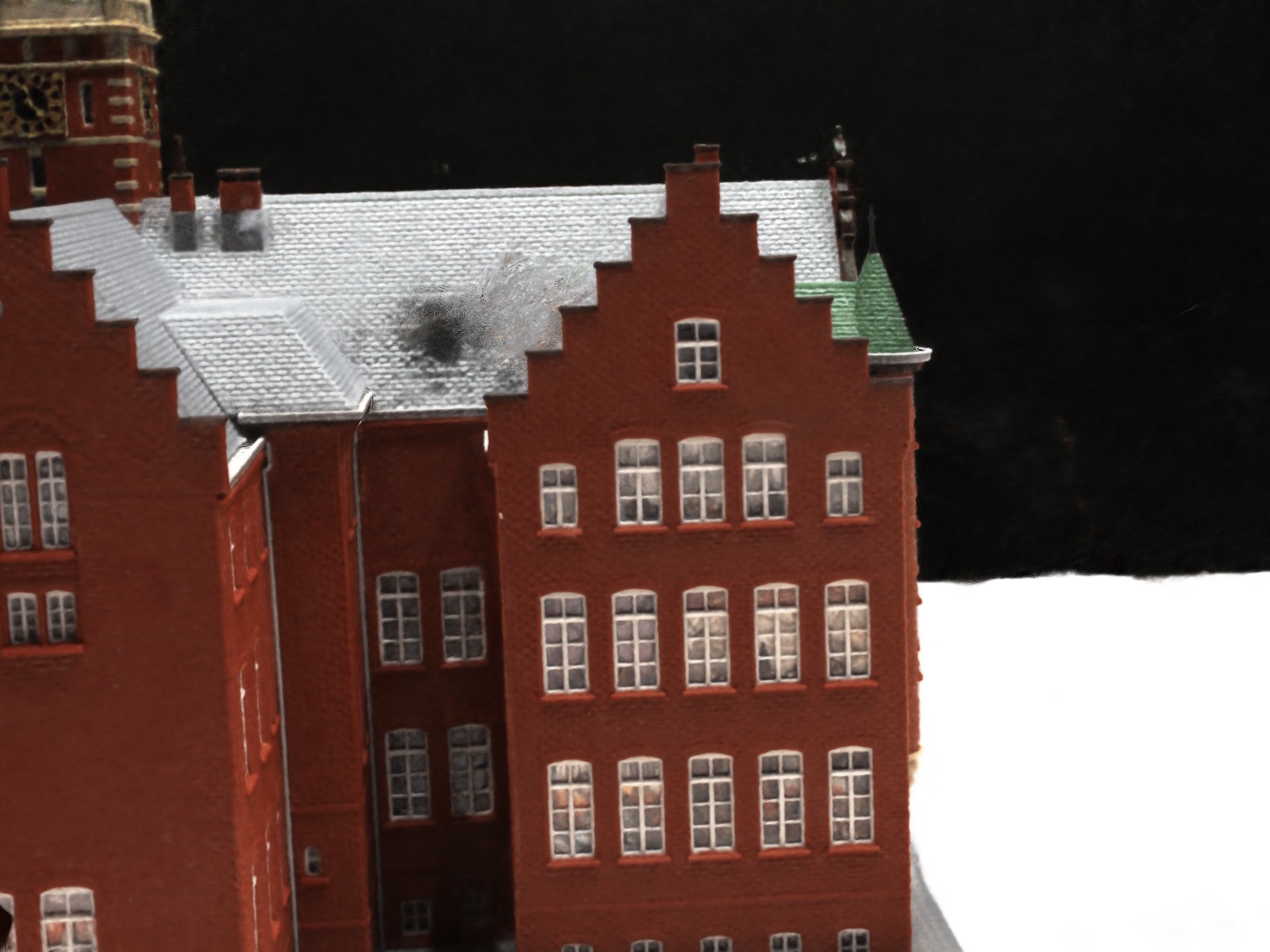} &
        \includegraphics[width=0.162\linewidth]{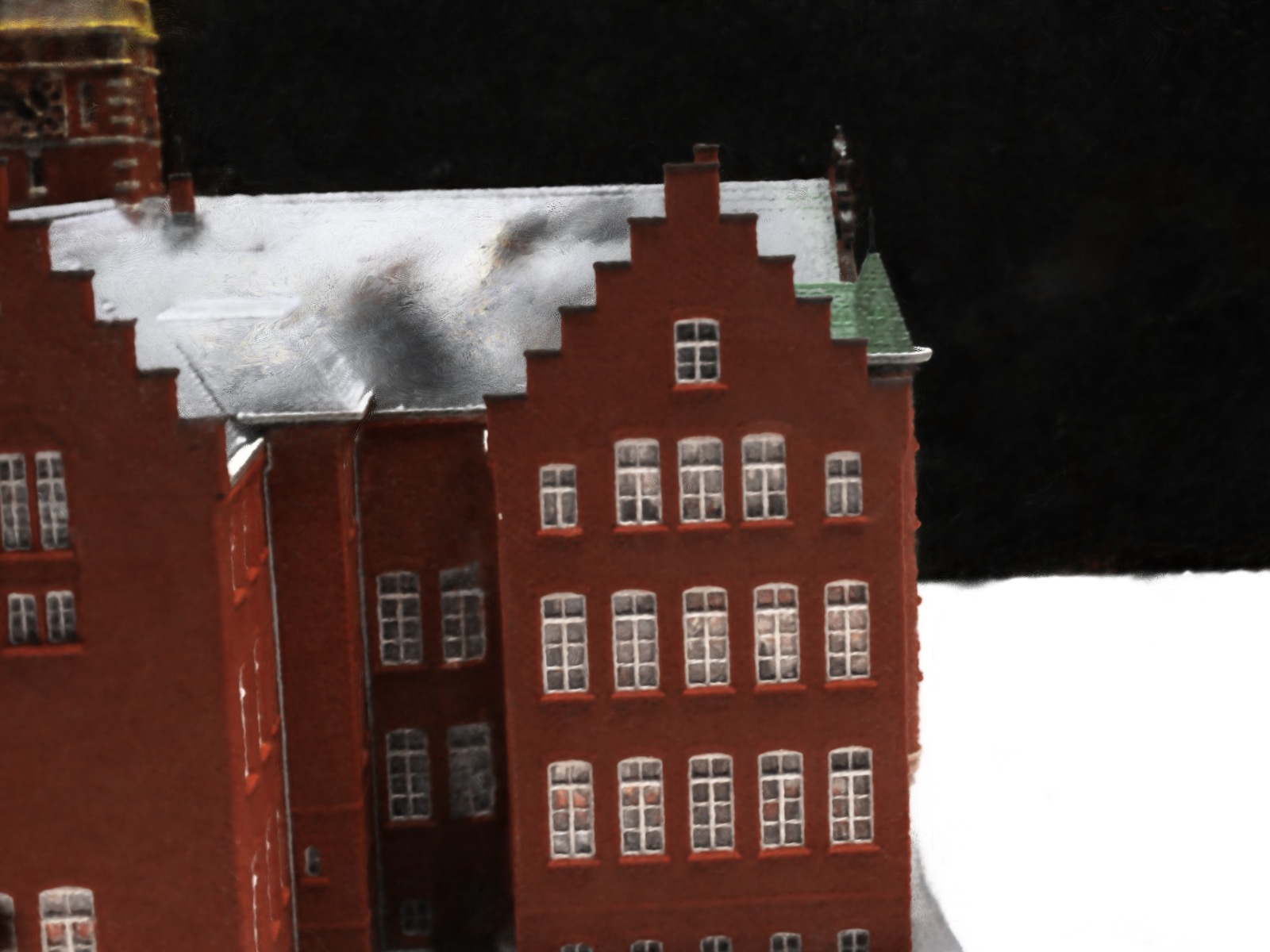} &
        \includegraphics[width=0.162\linewidth]{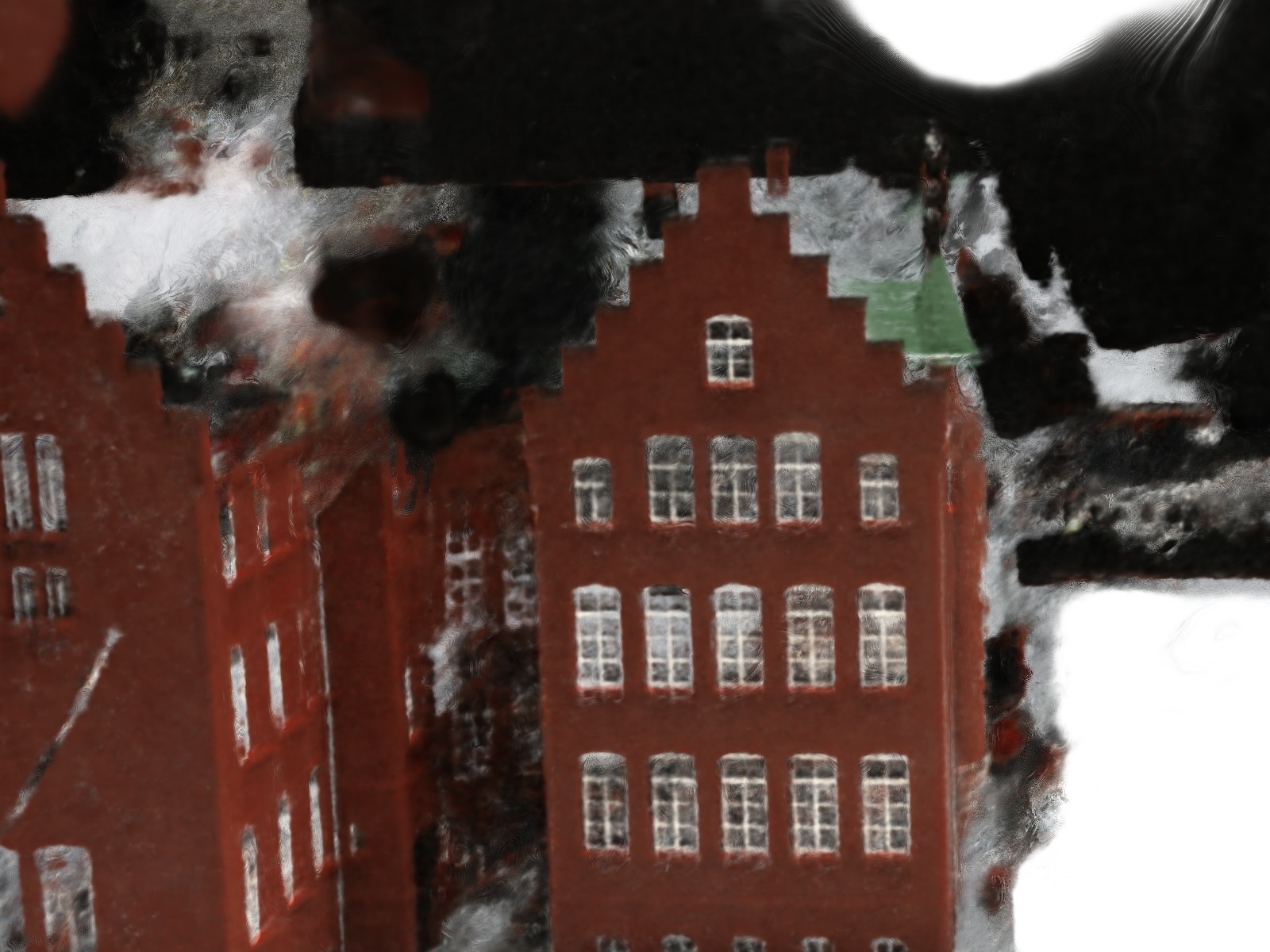} \\
    \end{tabular}
    \caption{Qualitative comparison under the 3-view protocol. Columns left to right: ground truth, DriftNeRF (ours), DNGaussian, FreeNeRF, NeRF. DriftNeRF produces the cleanest silhouettes and surfaces; NeRF exhibits severe floaters; DNGaussian and FreeNeRF fall in between. Top: Lego (NeRF-Synthetic). Bottom: Scan~24 (DTU). \Cref{fig:teaser} shows DTU Scan~34 under the same protocol, and further scenes are in \cref{app:qualitative}.}
    \Description{A two-row by five-column grid of renderings for Lego and DTU Scan 24. Columns are ground truth, DriftNeRF, DNGaussian, FreeNeRF, and NeRF. The NeRF column has the most floaters and haze, DNGaussian and FreeNeRF have fewer, and the DriftNeRF column has the cleanest silhouettes and surfaces.}
    \label{fig:qualitative}
\end{figure*}

\Cref{fig:qualitative} shows that DriftNeRF removes many isolated floaters and reduces haze, though the differences are not equally strong on every scene.
The per-scene NeRF-Synthetic results show gains over FreeNeRF on six of eight scenes and small regressions on Materials and Ship.
Both contain specular or thin structures for which Euclidean proximity is a poor proxy for surface membership; more examples are in \cref{app:qualitative}.

\subsection{Ablations and Sensitivity}
\label{sec:exp_ablation}
\begin{table}[t]
\centering
\caption{Subtractive ablation on NeRF-Synthetic, 3 views (eight-scene mean). The full system is \colorbox{baselineGray}{shaded} as the reference row; each row below removes one component from it. The no-stop-gradient control differentiates through both anchor and candidate densities.}
\label{tab:ablation}
\setlength{\tabcolsep}{5pt}
\begin{tabular}{@{}lccc@{}}
\toprule
Variant & PSNR$\uparrow$ & SSIM$\uparrow$ & LPIPS$\downarrow$\\
\midrule
\rowcolor{baselineGray} Ours (full) & 20.83 & .768 & .274\\
$-$ warmup & 20.41 & .758 & .282\\
$-$ stop-gradient & 20.58 & .759 & .278\\
$-$ drifting loss (NeRF) & 15.83 & .612 & .421\\
\bottomrule
\end{tabular}
\end{table}

Removing warmup reduces PSNR by 0.42\,dB, supporting the choice to establish an initial field before enforcing local agreement.
Removing stop-gradient reduces PSNR by 0.25\,dB.
As argued in \cref{sec:method_drift}, this control is the bilateral density smoother that corresponds to our formulation, so the gap is the value of the frozen target over local smoothing rather than a generic regularization effect.
At 50k iterations, mean PSNR over Lego, Materials, and Ship is 16.42 for the exponential kernel, 16.07 for a Gaussian kernel, and 16.28 for a diagonal $\ell_1$ kernel, consistent with the heavier-tail argument in \cref{sec:method_drift}.

\begin{table}[t]
\centering
\caption{Hyperparameter sensitivity on Lego (3 views, 50k iterations). Default values are \colorbox{baselineGray}{shaded}. Extended SSIM/LPIPS and 200k-iteration results are in \cref{app:hyperparam}.}
\label{tab:hyperparam_sensitivity}
\setlength{\tabcolsep}{3pt}
\footnotesize
\begin{tabular}{@{}llccccc@{}}
\toprule
Param & Values & \multicolumn{5}{c}{PSNR$\uparrow$}\\
\midrule
$\lambda_{\mathrm{drift}}$ & 0.01 / 0.05 / \textbf{0.1} / 0.5 / 1.0
            & 19.24 & 19.53 & \cellcolor{baselineGray}19.71 & 19.78 & 19.41\\
$\tau$      & 0.01 / 0.05 / \textbf{0.1} / 0.2 / 0.5
            & 18.94 & 19.52 & \cellcolor{baselineGray}19.71 & 19.43 & 18.67\\
$d_{\max}$  & 0.1 / 0.2 / \textbf{0.5} / 1.0 / 2.0
            & 18.81 & 19.38 & \cellcolor{baselineGray}19.71 & 19.56 & 19.12\\
$P$         & 256 / 512 / \textbf{2048} / 4096 / 8192
            & 19.14 & 19.42 & \cellcolor{baselineGray}19.71 & 19.74 & 19.73\\
$Q$         & 64 / 128 / \textbf{256} / 512 / 1024
            & 19.31 & 19.58 & \cellcolor{baselineGray}19.71 & 19.69 & 19.67\\
\bottomrule
\end{tabular}
\end{table}

\Cref{tab:hyperparam_sensitivity} sweeps each hyperparameter of \cref{eq:total_loss} and the kernel around its default.
The two spatial scales dominate: the worst setting of $\tau$ costs $1.04$\,dB and the worst setting of $d_{\max}$ costs $0.90$\,dB relative to the default, whereas the two sample counts saturate, with every $P\geq2048$ and $Q\geq256$ within $0.04$\,dB of the best value in its own sweep.
The sweep also shows that $\lambda=0.5$ edges out the default on Lego, by $19.78$ against $19.71$\,dB.
We keep $\lambda=0.1$ because that $0.07$\,dB deficit on one scene buys stability on the failure-sensitive ones, where $\lambda=0.5$ trails on Materials, Ship, and Drums, and because we prefer a single global value to per-scene tuning.

\paragraph{Efficiency.}
On an A100, the recorded per-iteration totals are 129.4\,ms for the vanilla implementation and 138.8\,ms for DriftNeRF, a 7.3\% increase.
The loss creates a $P\times Q$ distance matrix and has $O(PQ)$ auxiliary memory; measured peak GPU memory rises from 6.2 to 6.8\,GB, a 9.7\% increase, at the default $P{=}2048$, $Q{=}256$, itemized in \cref{tab:timing}.

\subsection{Dense-View Behavior}
\label{sec:exp_dense}

The regularizer should be inert when the rendering loss is already well constrained, and it is: under the conventional dense splits DriftNeRF stays within $0.15$\,dB of vanilla NeRF, moving DTU from $28.43$ to $28.57$\,dB and NeRF-Synthetic from $31.02$ to $31.09$\,dB, with SSIM and LPIPS changing by at most $0.003$.
Instant-NGP leads both dense benchmarks by a wide margin ($29.81$ and $33.18$\,dB), which reflects its hash-grid backbone rather than any regularization strategy and is consistent with \cref{tab:backbone}; \cref{tab:dense_results} gives the full table.

\section{Conclusion}
\label{sec:conclusion}

The per-ray rendering loss leaves cross-ray density agreement unconstrained, and the resulting disagreement is not a figure of speech: it can be measured during training on any density field, it persists in models that fit their training pixels, and it moves together with held-out quality when view coverage is thin.
DriftNeRF closes the gap by instantiating the drifting-field template on cross-ray density samples, where a stop-gradient consensus target supplies the asymmetry that separates the objective from local smoothing.
Doing so reduces residual disagreement on NeRF-Synthetic by an order of magnitude and improves 3-view PSNR by up to $+0.65$\,dB over the strongest MLP-NeRF baseline, at ${\sim}7\%$ overhead per iteration and with no external priors.
Because the signal is read off densities rather than architecture, it transfers to a hash-grid backbone unchanged, and under a matched 3-view protocol it holds up against recent sparse-view Gaussian pipelines without their depth, flow, or covisibility preprocessing, though at a substantially larger per-scene optimization cost.

\paragraph{Scope and extensions.}
The current implementation deliberately uses a simple isotropic Euclidean kernel, which keeps the regularizer independent of auxiliary geometry.
The Materials and Ship regressions suggest the next opportunity: incorporate local orientation, visibility, or manifold cues so that thin and specular structures receive more selective neighborhoods.
The same agreement principle may also transfer to explicit Gaussian primitives, whose positions and opacities provide natural candidates for cross-primitive consensus.

\section*{Reproducibility and Use of Generative AI}
Code for the drifting loss, the dataset loaders behind \cref{tab:dataset_protocol}, and the training, evaluation, and rendering entry points is at \url{https://anonymous.4open.science/r/DriftNeRF-DCF3/}; per-dataset hyperparameters are in \cref{tab:training_config}.
Generative AI was used for prose copy-editing and LaTeX formatting only; all code, algorithms, evaluation pipelines, and results were designed, implemented, executed, and verified by the authors.

\bibliographystyle{ACM-Reference-Format}
\bibliography{ref,refs-pending}

\appendix
\crefalias{section}{appendix}
\crefalias{subsection}{subappendix}

\section{Per-Scene Results}
\label{app:perscene}

Throughout the appendix, \colorbox{tabfirst}{first}, \colorbox{tabsecond}{second}, and \colorbox{tabthird}{third} rank per column among the three compared methods; tied values share the higher rank and the next colour is skipped.
Tables in the main text mark best values in bold instead, because they compare a longer method list.
The main text reports dataset-averaged metrics (\cref{tab:sparse_results}).
Below we break down PSNR, SSIM, and LPIPS per scene under the 3-view protocol for the three strongest methods: FreeNeRF, DNGaussian, and DriftNeRF (Ours).
We omit NeRF from per-scene tables as it is uniformly outperformed by all regularized methods.

\subsection{DTU (15 Scans)}

\begin{table}[ht]
    \centering
    \caption{Per-scene PSNR$\uparrow$ on DTU under the 3-view protocol. DriftNeRF improves over FreeNeRF on 13 of 15 scans; FreeNeRF retains an edge on Scans 41 and 103, and DNGaussian achieves the top score on Scan 31.}
    \label{tab:perscene_dtu_psnr}
    \setlength{\tabcolsep}{0pt}
    \scriptsize
    \begin{tabular}{@{}l@{\hskip 3pt}ccccccccccccccc|c@{}}
        \toprule
        Method & 8 & 21 & 30 & 31 & 34 & 38 & 40 & 41 & 45 & 55 & 63 & 82 & 103 & 110 & 114 & Avg \\
        \midrule
        DNGaussian & \second{19.07} & \second{17.56} & \second{17.84} & \first{19.62} & \third{18.34} & \third{16.73} & \third{17.98} & \third{15.81} & \third{17.82} & \second{20.08} & \second{20.81} & \third{18.49} & \third{16.32} & \third{15.67} & \third{17.26} & \second{17.96} \\
        FreeNeRF & \third{18.71} & \third{17.24} & \third{17.58} & \third{19.03} & \second{18.42} & \second{16.89} & \second{18.17} & \first{16.23} & \second{17.94} & \third{19.37} & \third{20.12} & \second{18.63} & \first{16.78} & \second{15.94} & \second{17.30} & \third{17.89} \\
        Ours     & \first{19.14} & \first{17.68} & \first{18.03} & \second{19.47} & \first{18.89} & \first{17.21} & \first{18.62} & \second{16.08} & \first{18.37} & \first{20.34} & \first{21.07} & \first{19.06} & \second{16.62} & \first{16.42} & \first{17.65} & \first{18.31} \\
        \bottomrule
    \end{tabular}
\end{table}

\begin{table}[ht]
    \centering
    \caption{Per-scene SSIM$\uparrow$ / LPIPS$\downarrow$ on DTU under the 3-view protocol.}
    \label{tab:perscene_dtu_ssim_lpips}
    \setlength{\tabcolsep}{1.2pt}
    \scriptsize
    \begin{tabular}{@{}l*{15}{c}|c@{}}
        \toprule
        \multicolumn{17}{c}{\textbf{SSIM}$\uparrow$} \\
        \midrule
        Method & 8 & 21 & 30 & 31 & 34 & 38 & 40 & 41 & 45 & 55 & 63 & 82 & 103 & 110 & 114 & Avg \\
        \midrule
        DNGaussian & \second{.731} & \second{.673} & \second{.687} & \second{.744} & \second{.710} & \second{.656} & \second{.700} & \third{.623} & \second{.695} & \second{.749} & \second{.765} & \second{.712} & \third{.638} & \third{.614} & \second{.713} & \second{.694} \\
        FreeNeRF & \third{.724} & \third{.667} & \third{.681} & \third{.738} & \third{.703} & \third{.652} & \third{.694} & \first{.631} & \third{.689} & \third{.742} & \third{.758} & \third{.706} & \first{.645} & \second{.617} & \third{.688} & \third{.689} \\
        Ours     & \first{.732} & \first{.675} & \first{.689} & \first{.746} & \first{.712} & \first{.659} & \first{.703} & \second{.626} & \first{.697} & \first{.751} & \first{.768} & \first{.714} & \second{.641} & \first{.628} & \first{.714} & \first{.697} \\
        \midrule
        \multicolumn{17}{c}{\textbf{LPIPS}$\downarrow$} \\
        \midrule
        Method & 8 & 21 & 30 & 31 & 34 & 38 & 40 & 41 & 45 & 55 & 63 & 82 & 103 & 110 & 114 & Avg \\
        \midrule
        DNGaussian & \first{.278} & \first{.332} & \first{.318} & \first{.264} & \first{.298} & \first{.347} & \first{.308} & \second{.386} & \first{.312} & \first{.259} & \first{.246} & \first{.296} & \second{.368} & \second{.386} & \first{.267} & \first{.311} \\
        FreeNeRF & \third{.296} & \third{.347} & \third{.334} & \third{.283} & \third{.317} & \third{.361} & \third{.324} & \first{.383} & \third{.329} & \third{.278} & \third{.264} & \third{.312} & \first{.367} & \third{.398} & \third{.327} & \third{.328} \\
        Ours     & \second{.286} & \second{.336} & \second{.324} & \second{.273} & \second{.306} & \second{.350} & \second{.313} & \third{.394} & \second{.317} & \second{.268} & \second{.252} & \second{.301} & \third{.377} & \first{.385} & \second{.273} & \second{.317} \\
        \bottomrule
    \end{tabular}
\end{table}

DriftNeRF improves on 13 of 15 DTU scans over FreeNeRF across all three metrics.
FreeNeRF retains an edge on Scans~41 and 103, which contain fine geometric detail and reflective surfaces where the isotropic kernel conflates nearby densities across narrow structures, consistent with the Materials/Ship pattern observed on NeRF-Synthetic.
DNGaussian achieves the top PSNR on Scan~31 and outperforms Ours on per-scene LPIPS on 14 of 15 scans, a cross-metric disagreement indicating that explicit Gaussian primitives produce perceptually sharper textures even when overall PSNR is lower.

\subsection{LLFF (8 Scenes)}

\begin{table}[ht]
    \centering
    \caption{Per-scene PSNR$\uparrow$ on LLFF under the 3-view protocol. DriftNeRF improves on 6 of 8 scenes; FreeNeRF retains an edge on Orchids and Flower. DNGaussian struggles on thin organic structures (Orchids, Flower).}
    \label{tab:perscene_llff_psnr}
    \scriptsize
    \setlength{\tabcolsep}{2.6pt}
    \begin{tabular}{@{}lcccccccc|c@{}}
        \toprule
        Method & Room & Fern & Leaves & Fortress & Orchids & Flower & T-Rex & Horns & Avg \\
        \midrule
        DNGaussian & \second{22.17} & \third{18.78} & \third{16.89} & \second{22.84} & \third{16.21} & \third{17.94} & \third{20.63} & \second{19.10} & \third{19.32} \\
        FreeNeRF & \third{21.83} & \second{19.12} & \second{17.34} & \third{22.56} & \first{16.74} & \first{18.47} & \second{21.08} & \third{18.14} & \second{19.41} \\
        Ours     & \first{22.31} & \first{19.58} & \first{17.82} & \first{22.94} & \second{16.53} & \second{18.24} & \first{21.57} & \first{19.65} & \first{19.83} \\
        \bottomrule
    \end{tabular}
\end{table}

\begin{table}[ht]
    \centering
    \caption{Per-scene SSIM$\uparrow$ / LPIPS$\downarrow$ on LLFF under the 3-view protocol.}
    \label{tab:perscene_llff_ssim_lpips}
    \scriptsize
    \setlength{\tabcolsep}{2.6pt}
    \begin{tabular}{@{}lcccccccc|c@{}}
        \toprule
        \multicolumn{10}{c}{\textbf{SSIM}$\uparrow$} \\
        \midrule
        Method & Room & Fern & Leaves & Fortress & Orchids & Flower & T-Rex & Horns & Avg \\
        \midrule
        DNGaussian & \second{.714} & \third{.604} & \third{.531} & \second{.735} & \third{.509} & \third{.574} & \third{.664} & \third{.589} & \third{.615} \\
        FreeNeRF & \third{.704} & \second{.612} & \second{.543} & \third{.726} & \first{.524} & \first{.589} & \second{.672} & \second{.598} & \second{.621} \\
        Ours     & \first{.724} & \first{.631} & \first{.558} & \first{.743} & \second{.517} & \second{.581} & \first{.691} & \first{.659} & \first{.638} \\
        \midrule
        \multicolumn{10}{c}{\textbf{LPIPS}$\downarrow$} \\
        \midrule
        Method & Room & Fern & Leaves & Fortress & Orchids & Flower & T-Rex & Horns & Avg \\
        \midrule
        DNGaussian & \second{.286} & \third{.382} & \third{.447} & \second{.262} & \third{.464} & \third{.401} & \third{.329} & \third{.421} & \third{.374} \\
        FreeNeRF & \third{.298} & \second{.374} & \second{.437} & \third{.268} & \first{.452} & \first{.392} & \second{.321} & \second{.402} & \second{.368} \\
        Ours     & \first{.281} & \first{.358} & \first{.421} & \first{.253} & \second{.461} & \second{.398} & \first{.304} & \first{.340} & \first{.352} \\
        \bottomrule
    \end{tabular}
\end{table}

DriftNeRF improves on 6 of 8 LLFF scenes.
FreeNeRF retains an edge on Orchids and Flower, both of which contain fine petal and leaf detail at varying depths; the isotropic kernel pools densities across thin overlapping structures, consistent with the thin-structure limitation discussed in \cref{sec:exp_sparse}.
DNGaussian consistently places third on scenes with fine organic structures, where explicit Gaussian primitives are less effective than frequency-based or cross-ray regularization.

\subsection{NeRF-Synthetic (8 Scenes)}

\begin{table}[ht]
    \centering
    \caption{Per-scene PSNR$\uparrow$ on NeRF-Synthetic under the 3-view protocol. These eight scene means are the input to the paired Wilcoxon test reported in \cref{sec:exp_sparse}. DriftNeRF improves over FreeNeRF on 6 of 8 scenes and holds the top score among the three methods on 5 of 8; FreeNeRF retains an edge on Materials and Ship.}
    \label{tab:perscene_synthetic_psnr}
    \scriptsize
    \setlength{\tabcolsep}{2.6pt}
    \begin{tabular}{@{}lcccccccc|c@{}}
        \toprule
        Method & Chair & Drums & Ficus & Hotdog & Lego & Matrl. & Mic & Ship & Avg \\
        \midrule
        DNGaussian & \first{25.44} & \second{17.21} & \third{20.01} & \third{22.96} & \second{22.34} & \third{18.38} & \second{22.52} & \third{13.62} & \second{20.31} \\
        FreeNeRF & \third{24.31} & \third{16.83} & \second{20.42} & \second{23.17} & \third{21.64} & \first{18.92} & \third{22.08} & \first{14.07} & \third{20.18} \\
        Ours     & \second{25.38} & \first{17.59} & \first{21.24} & \first{24.06} & \first{22.78} & \second{18.76} & \first{22.85} & \second{13.98} & \first{20.83} \\
        \bottomrule
    \end{tabular}
\end{table}

\begin{table}[ht]
    \centering
    \caption{Per-scene SSIM$\uparrow$ / LPIPS$\downarrow$ on NeRF-Synthetic under the 3-view protocol.}
    \label{tab:perscene_synthetic_ssim_lpips}
    \scriptsize
    \setlength{\tabcolsep}{2.6pt}
    \begin{tabular}{@{}lcccccccc|c@{}}
        \toprule
        \multicolumn{10}{c}{\textbf{SSIM}$\uparrow$} \\
        \midrule
        Method & Chair & Drums & Ficus & Hotdog & Lego & Matrl. & Mic & Ship & Avg \\
        \midrule
        DNGaussian & \first{.871} & \second{.656} & \third{.770} & \second{.829} & \second{.794} & \second{.704} & \second{.812} & \third{.652} & \second{.761} \\
        FreeNeRF & \third{.842} & \third{.641} & \second{.774} & \third{.828} & \third{.771} & \first{.718} & \third{.796} & \first{.678} & \third{.756} \\
        Ours     & \second{.864} & \first{.662} & \first{.791} & \first{.843} & \first{.801} & \second{.704} & \first{.822} & \second{.657} & \first{.768} \\
        \midrule
        \multicolumn{10}{c}{\textbf{LPIPS}$\downarrow$} \\
        \midrule
        Method & Chair & Drums & Ficus & Hotdog & Lego & Matrl. & Mic & Ship & Avg \\
        \midrule
        DNGaussian & \first{.142} & \first{.376} & \third{.250} & \third{.189} & \second{.228} & \third{.334} & \second{.208} & \third{.521} & \second{.281} \\
        FreeNeRF & \third{.168} & \third{.413} & \second{.248} & \second{.184} & \third{.254} & \first{.312} & \third{.230} & \first{.487} & \third{.287} \\
        Ours     & \second{.148} & \second{.384} & \first{.241} & \first{.172} & \first{.218} & \second{.326} & \first{.196} & \second{.507} & \first{.274} \\
        \bottomrule
    \end{tabular}
\end{table}

Against FreeNeRF alone, DriftNeRF improves PSNR on 6 of 8 NeRF-Synthetic scenes, matching the count reported in \cref{sec:exp_sparse}.
Among the three methods compared here, it holds the top PSNR and SSIM on 5 of 8 scenes and the top LPIPS on 4 of 8, with Chair going to DNGaussian on all three metrics.
FreeNeRF retains an edge on Materials and Ship, both of which contain thin structures (Ship's rigging) and specular surfaces (Materials' reflective sphere) where the isotropic kernel conflates densities across narrow gaps.
DNGaussian achieves the top score on Chair across all three metrics.
On Ship, all methods yield high LPIPS ($>0.48$), reflecting the intrinsic difficulty of reconstructing fine rigging geometry from 3 views.

\section{Dense-View Results}
\label{app:dense}

\Cref{sec:exp_dense} summarizes the dense-view control in prose; \cref{tab:dense_results} gives the full table.
Under the conventional dense splits the drifting loss is inert, as it should be when the rendering loss is already well constrained: DriftNeRF stays within $0.15$~dB of vanilla NeRF on both benchmarks, and SSIM and LPIPS change by at most $0.003$.
Instant-NGP leads both benchmarks by a wide margin, which reflects its hash-grid backbone rather than any regularization strategy.
LLFF has no separate dense regime under this protocol and is omitted.

\begin{table}[ht]
    \centering
    \caption{Dense-view results on DTU and NeRF-Synthetic under the conventional splits.}
    \label{tab:dense_results}
    \small
    \setlength{\tabcolsep}{4.5pt}
    \begin{tabular}{@{}lcccccc@{}}
        \toprule
        & \multicolumn{3}{c}{DTU} & \multicolumn{3}{c}{NeRF-Synthetic}\\
        \cmidrule(lr){2-4}\cmidrule(lr){5-7}
        Method & PSNR$\uparrow$ & SSIM$\uparrow$ & LPIPS$\downarrow$ & PSNR$\uparrow$ & SSIM$\uparrow$ & LPIPS$\downarrow$\\
        \midrule
        NeRF        & \third{28.43} & \third{.917} & \third{.189} & \third{31.02} & \third{.947} & \third{.058}\\
        Instant-NGP & \first{29.81} & \first{.931} & \first{.162} & \first{33.18} & \first{.962} & \first{.038}\\
        Ours        & \second{28.57} & \second{.918} & \second{.186} & \second{31.09} & \second{.948} & \second{.057}\\
        \bottomrule
    \end{tabular}
\end{table}

\section{Extended Hyperparameter Sensitivity}
\label{app:hyperparam}

\Cref{tab:hyperparam_sensitivity} reports PSNR sensitivity to each drifting field hyperparameter on Lego (3 views, 50k iterations).
\Cref{tab:hyperparam_full} below extends that analysis with SSIM and LPIPS.
The conclusions are consistent across all metrics: performance is robust for $\tau \in [0.05, 0.2]$, $d_{\max} \in [0.2, 1.0]$, and $P \geq 2048$.

\begin{table}[ht]
    \centering
    \caption{Extended hyperparameter sensitivity on Lego (3 views, 50k iters) with SSIM$\uparrow$ and LPIPS$\downarrow$. Default values are \colorbox{baselineGray}{shaded}.}
    \label{tab:hyperparam_full}
    \setlength{\tabcolsep}{2pt}
    \scriptsize
    \begin{tabular}{@{}ll*{5}{c}*{5}{c}@{}}
        \toprule
        & & \multicolumn{5}{c}{SSIM$\uparrow$} & \multicolumn{5}{c}{LPIPS$\downarrow$} \\
        \midrule
        $\reg_{\mathrm{drift}}$ & 0.01 / 0.05 / \textbf{0.1} / 0.5 / 1.0
            & .742 & .754 & \cellcolor{baselineGray}.761 & .764 & .748
            & .298 & .286 & \cellcolor{baselineGray}.278 & .275 & .293 \\
        $\tau$ & 0.01 / 0.05 / \textbf{0.1} / 0.2 / 0.5
            & .731 & .753 & \cellcolor{baselineGray}.761 & .749 & .718
            & .312 & .288 & \cellcolor{baselineGray}.278 & .291 & .324 \\
        $d_{\max}$ & 0.1 / 0.2 / \textbf{0.5} / 1.0 / 2.0
            & .724 & .746 & \cellcolor{baselineGray}.761 & .756 & .738
            & .318 & .294 & \cellcolor{baselineGray}.278 & .284 & .302 \\
        $P$ & 256 / 512 / \textbf{2048} / 4096 / 8192
            & .739 & .751 & \cellcolor{baselineGray}.761 & .762 & .761
            & .302 & .289 & \cellcolor{baselineGray}.278 & .277 & .277 \\
        $Q$ & 64 / 128 / \textbf{256} / 512 / 1024
            & .747 & .756 & \cellcolor{baselineGray}.761 & .760 & .758
            & .293 & .284 & \cellcolor{baselineGray}.278 & .279 & .281 \\
        \bottomrule
    \end{tabular}
\end{table}

\Cref{tab:hyperparam_sensitivity} reports sensitivity at 50k iterations.
\Cref{tab:hyperparam_200k} extends this to full 200k-iteration training.
The same sensitivity patterns persist at convergence: $\tau$ and $d_{\max}$ remain the most sensitive parameters, while $P$ and $Q$ saturate early.

\begin{table}[ht]
    \centering
    \caption{Hyperparameter sensitivity on Lego (3 views, 200k iters) with PSNR$\uparrow$. Default values are \colorbox{baselineGray}{shaded}. Compare with \cref{tab:hyperparam_sensitivity} (50k).}
    \label{tab:hyperparam_200k}
    \small
    \setlength{\tabcolsep}{3pt}
    \begin{tabular}{@{}llccccc@{}}
        \toprule
        Param & Values & \multicolumn{5}{c}{PSNR$\uparrow$} \\
        \midrule
        $\reg_{\mathrm{drift}}$ & 0.01 / 0.05 / \textbf{0.1} / 0.5 / 1.0
            & 21.72 & 22.34 & \cellcolor{baselineGray}22.78 & 22.91 & 21.83 \\
        $\tau$ & 0.01 / 0.05 / \textbf{0.1} / 0.2 / 0.5
            & 20.86 & 22.28 & \cellcolor{baselineGray}22.78 & 22.17 & 20.24 \\
        $d_{\max}$ & 0.1 / 0.2 / \textbf{0.5} / 1.0 / 2.0
            & 20.63 & 22.08 & \cellcolor{baselineGray}22.78 & 22.39 & 21.46 \\
        $P$ & 256 / 512 / \textbf{2048} / 4096 / 8192
            & 21.62 & 22.24 & \cellcolor{baselineGray}22.78 & 22.83 & 22.81 \\
        $Q$ & 64 / 128 / \textbf{256} / 512 / 1024
            & 22.01 & 22.46 & \cellcolor{baselineGray}22.78 & 22.76 & 22.73 \\
        \bottomrule
    \end{tabular}
\end{table}

\section{Implementation Details}
\label{app:impl}

\subsection{Drifting Loss Pseudocode}

\Cref{alg:drift} details the drifting loss computation: subsample anchor and candidate points, compute pairwise distances, mask same-ray pairs and pairs beyond $d_{\max}$, evaluate the kernel-weighted consensus, and compute the MSE with the stop-gradient target.

\begin{algorithm}[ht]
\caption{Drifting Loss Computation}
\label{alg:drift}
\begin{algorithmic}[1]
\REQUIRE Sample points $\mathbf{x} \in \mathbb{R}^{N \times S \times 3}$, densities $\boldsymbol{\sigma} \in \mathbb{R}^{N \times S}$, ray indices $\mathbf{r} \in \{1,\dots,N\}^{N \times S}$
\REQUIRE Hyperparameters: $P$ (anchors), $Q$ (candidates), $\tau$ (temperature), $d_{\max}$ (cutoff)
\STATE Flatten: $\mathbf{x}_{\mathrm{flat}} \in \mathbb{R}^{(NS) \times 3}$,\; $\boldsymbol{\sigma}_{\mathrm{flat}} \in \mathbb{R}^{NS}$,\; $\mathbf{r}_{\mathrm{flat}} \in \mathbb{R}^{NS}$
\STATE Sample $P$ anchor indices $\mathcal{A} \subset \{1,\dots,NS\}$ uniformly
\STATE Sample $Q$ candidate indices $\mathcal{C} \subset \{1,\dots,NS\}$ uniformly
\STATE $\mathbf{D} \leftarrow \texttt{cdist}(\mathbf{x}_{\mathrm{flat}}[\mathcal{A}],\; \mathbf{x}_{\mathrm{flat}}[\mathcal{C}])$ \COMMENT{$P \times Q$ distance matrix}
\STATE $\mathbf{K} \leftarrow \exp(-\mathbf{D} / \tau)$ \COMMENT{Exponential kernel}
\STATE $\mathbf{M}_{\mathrm{ray}} \leftarrow (\mathbf{r}_{\mathrm{flat}}[\mathcal{A}] \neq \mathbf{r}_{\mathrm{flat}}[\mathcal{C}]^\top)$ \COMMENT{cross-ray mask}
\STATE $\mathbf{M}_{\mathrm{dist}} \leftarrow (\mathbf{D} \leq d_{\max})$ \COMMENT{distance cutoff mask}
\STATE $\mathbf{M} \leftarrow \mathbf{M}_{\mathrm{ray}} \odot \mathbf{M}_{\mathrm{dist}}$
\STATE $\mathbf{W} \leftarrow \mathbf{K} \odot \mathbf{M}$ \COMMENT{masked kernel weights}
\STATE $\mathbf{W} \leftarrow \mathbf{W} \,/\, (\mathbf{W}.\text{sum}(\text{dim}=1, \text{keepdim}=\text{True}) + \epsilon)$ \COMMENT{normalize per anchor}
\STATE $\bar{\boldsymbol{\sigma}}_{\mathrm{cross}} \leftarrow \texttt{sg}(\mathbf{W} \cdot \boldsymbol{\sigma}_{\mathrm{flat}}[\mathcal{C}])$ \COMMENT{consensus target, stop-gradient}
\STATE $\Ls_{\mathrm{drift}} \leftarrow \frac{1}{P}\sum_{i \in \mathcal{A}} (\boldsymbol{\sigma}_{\mathrm{flat}}[i] - \bar{\boldsymbol{\sigma}}_{\mathrm{cross}}[i])^2$
\RETURN $\Ls_{\mathrm{drift}}$
\end{algorithmic}
\end{algorithm}

The stop-gradient operator $\sg(\cdot)$ on line~11 ensures that gradients flow only through the anchor density $\sigma_{a,i}$, implementing the drifting dynamics described in \cref{sec:method_drift}: each density \emph{drifts} in the direction of the drifting field~$V$, toward cross-ray consensus, with per-step update $\sigma_{a,i} \leftarrow \sigma_{a,i} + 2\alpha V$ (\cref{eq:drifting_gradient}).
The numerical stability constant $\epsilon = 10^{-8}$ prevents division by zero when an anchor has no valid cross-ray neighbors within $d_{\max}$.
Without stop-gradient, both anchor and candidate densities would receive competing gradients that encourage uniform density rather than consensus at the correct values, reducing PSNR by $0.25$~dB (\cref{tab:ablation}).

\subsection{Training Hyperparameters per Dataset}

\Cref{tab:training_config} summarizes all training hyperparameters for each benchmark.
All settings follow the vanilla NeRF defaults except where noted.

\begin{table}[ht]
    \centering
    \caption{Training configuration per dataset. All experiments use the Adam optimizer with $\beta_1=0.9$, $\beta_2=0.999$. No NDC transform is applied on any dataset; kernel distances are evaluated in loader coordinates.}
    \label{tab:training_config}
    \small
    \setlength{\tabcolsep}{3pt}
    \begin{tabular}{@{}lccc@{}}
        \toprule
        Parameter & NeRF-Syn. & DTU & LLFF \\
        \midrule
        Resolution & $800^2$ & $400{\times}300$ & $504{\times}378$ \\
        Near / far & 2.0 / 6.0 & scan cameras & \texttt{poses\_bounds} \\
        \midrule
        \multicolumn{4}{l}{\textit{Shared across datasets}} \\
        \midrule
        Coarse / fine samples & \multicolumn{3}{c}{64 / 64} \\
        Batch size (rays) & \multicolumn{3}{c}{1024} \\
        Iterations & \multicolumn{3}{c}{200k} \\
        Learning rate & \multicolumn{3}{c}{$5\times10^{-4} \to 5\times10^{-5}$} \\
        LR schedule & \multicolumn{3}{c}{exponential decay} \\
        Density activation & \multicolumn{3}{c}{softplus} \\
        Pos.\ encoding (xyz / dir) & \multicolumn{3}{c}{10 / 4 octaves} \\
        MLP depth $\times$ width & \multicolumn{3}{c}{$8 \times 256$} \\
        \midrule
        \multicolumn{4}{l}{\textit{Drifting field parameters}} \\
        \midrule
        $\reg_{\mathrm{drift}}$ & \multicolumn{3}{c}{0.1} \\
        Warmup & \multicolumn{3}{c}{linear over first 10\% of iterations} \\
        $\tau$ & \multicolumn{3}{c}{0.1} \\
        $d_{\max}$ & \multicolumn{3}{c}{0.5} \\
        $P$ (anchor points) & \multicolumn{3}{c}{2048} \\
        $Q$ (candidate points) & \multicolumn{3}{c}{256} \\
        \bottomrule
    \end{tabular}
\end{table}

\subsection{Wall-Clock Timing}

\Cref{tab:timing} reports per-iteration wall-clock timing on a single NVIDIA A100 GPU.
The drifting loss adds a modest overhead, primarily from the pairwise distance computation (\texttt{cdist}) and kernel evaluation.

\begin{table}[ht]
    \centering
    \caption{Per-iteration timing breakdown (ms/iter, A100 GPU). The drifting loss adds ${\sim}7\%$ wall-clock overhead.}
    \label{tab:timing}
    \begin{tabular}{@{}lcc@{}}
        \toprule
        Component & NeRF (ms) & DriftNeRF (ms) \\
        \midrule
        Forward pass (MLP) & 42.3 & 42.3 \\
        Volume rendering & 8.7 & 8.7 \\
        Rendering loss + backward & 78.4 & 78.4 \\
        Drifting loss (forward) & -- & 5.8 \\
        Drifting loss (backward) & -- & 3.6 \\
        \midrule
        Total & 129.4 & 138.8 \\
        Overhead & -- & +7.3\% \\
        \bottomrule
    \end{tabular}
\end{table}

Peak GPU memory usage is 6.2~GB for NeRF and 6.8~GB for DriftNeRF ($+$9.7\%), primarily due to the $P \times Q$ distance and kernel matrices stored for backpropagation.
Both figures are measured with the coarse and fine rendering losses retained, and match the totals quoted in \cref{sec:exp_ablation}.

\section{Reproduction Sheet for the Matched 3DGS Comparison}
\label{app:reprosheet}

Each baseline in the matched 3-view comparison (\cref{tab:threeview_3dgs}) is run from its official release, unmodified, on the exact camera indices, resolutions, and held-out-view evaluator of \cref{tab:dataset_protocol}.
\Cref{tab:reprosheet} records what each run consumed, so that a reader can tell which comparisons depend on auxiliary estimators rather than on the sparse images alone.

\begin{table}[ht]
    \centering
    \caption{Reproduction sheet for the matched 3-view 3DGS comparison. ``Priors'' lists auxiliary estimators applied to the three training images only.}
    \label{tab:reprosheet}
    \scriptsize
    \setlength{\tabcolsep}{3pt}
    \begin{tabular}{@{}llll@{}}
        \toprule
        Method & Release & Steps & Priors \\
        \midrule
        3DGS & \texttt{graphdeco/gaussian-splatting} & 30k & none (random init on NeRF-Syn.) \\
        FSGS & \texttt{VITA-Group/FSGS} & 10k & DPT monocular depth \\
        DNGaussian & \texttt{Fictionarry/DNGaussian} & 6k & monocular depth \\
        SparseGS & \texttt{ForMyCat/SparseGS} & 30k & monocular depth, floater pruning \\
        DropoutGS & official release & 30k & none beyond SfM \\
        NexusGS & \texttt{USMizuki/NexusGS} & 30k & optical flow, epipolar depth \\
        SE-GS & \texttt{sailor-z/SE-GS} & 30k & none beyond SfM \\
        \midrule
        Ours & this submission & 200k & none \\
        \bottomrule
    \end{tabular}
\end{table}

All runs use one NVIDIA A100 with CUDA 12.1 and PyTorch 2.5.1.
Structure-from-motion is run with COLMAP on the three training images per scene.
On NeRF-Synthetic, COLMAP returns no usable reconstruction from three white-background frames, so every Gaussian baseline that admits it falls back to the random point initialization shipped with 3DGS; this is the cause of the NeRF-Synthetic ordering discussed in \cref{sec:exp_sparse}.
CoMapGS is excluded because no public implementation is available, and SparseGS has no NeRF-Synthetic configuration in its release.

\section{Additional Qualitative Comparisons}
\label{app:qualitative}

\begin{figure*}[!t]
    \centering
    \setlength{\tabcolsep}{1pt}
    \begin{tabular}{cccccc}
        & GT & Ours & DNGaussian & FreeNeRF & NeRF \\[1pt]
        \rotatebox{90}{\small\;\;\;Ship} &
        \includegraphics[width=0.175\linewidth]{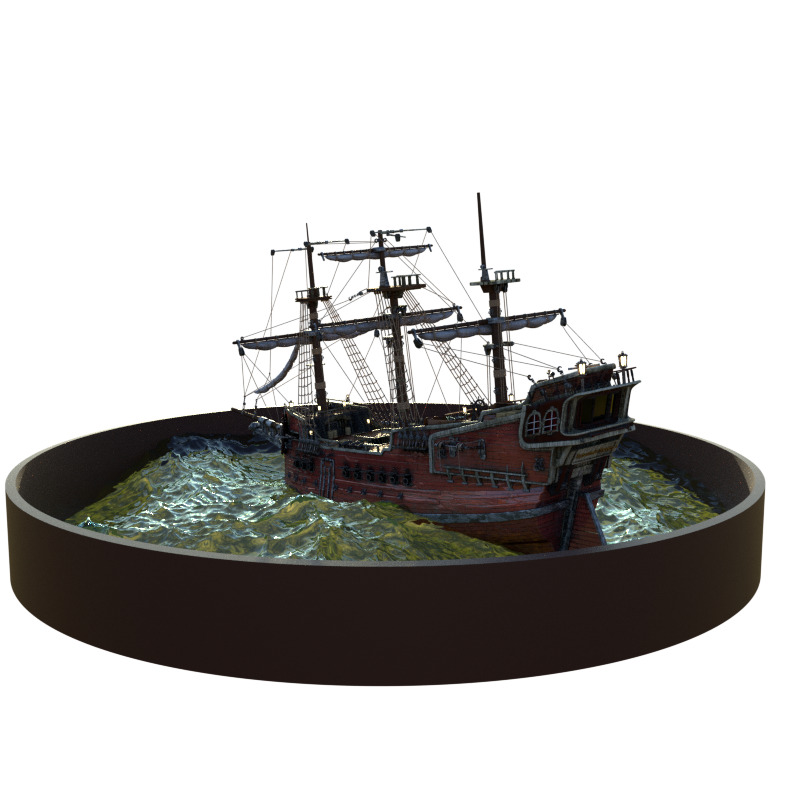} &
        \includegraphics[width=0.175\linewidth]{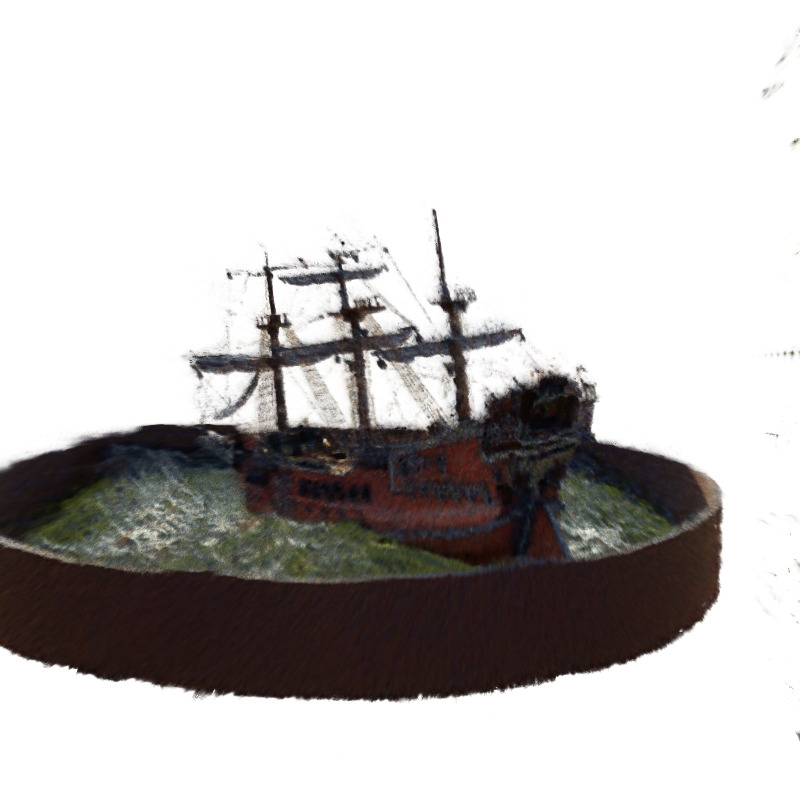} &
        \includegraphics[width=0.175\linewidth]{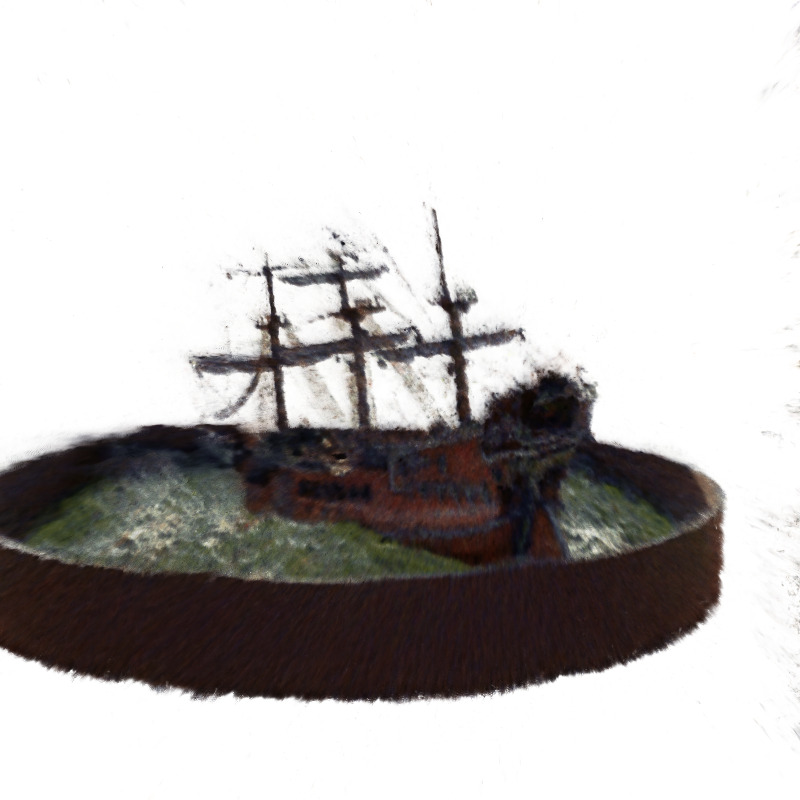} &
        \includegraphics[width=0.175\linewidth]{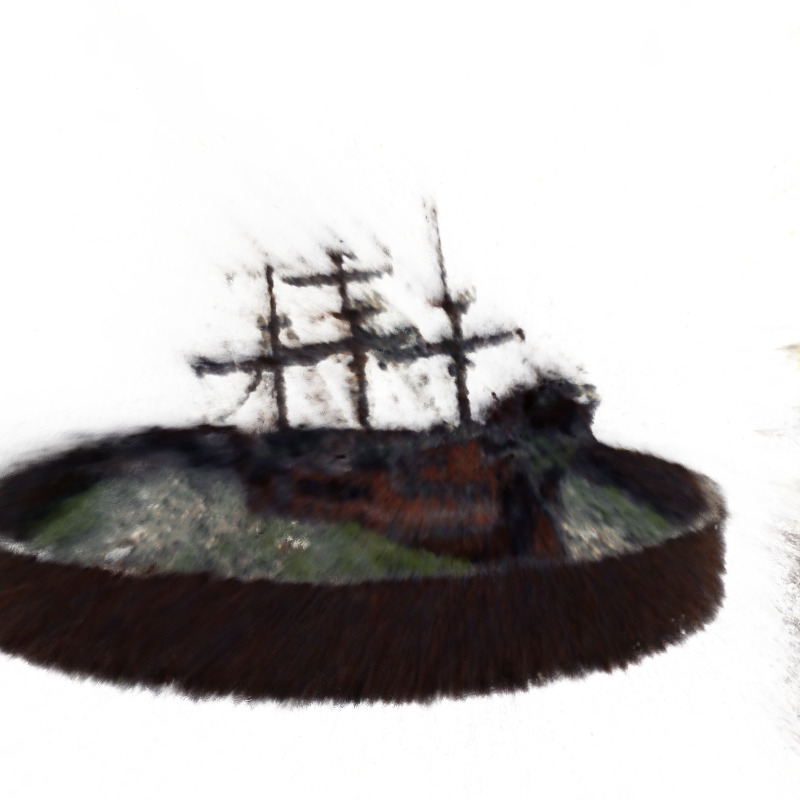} &
        \includegraphics[width=0.175\linewidth]{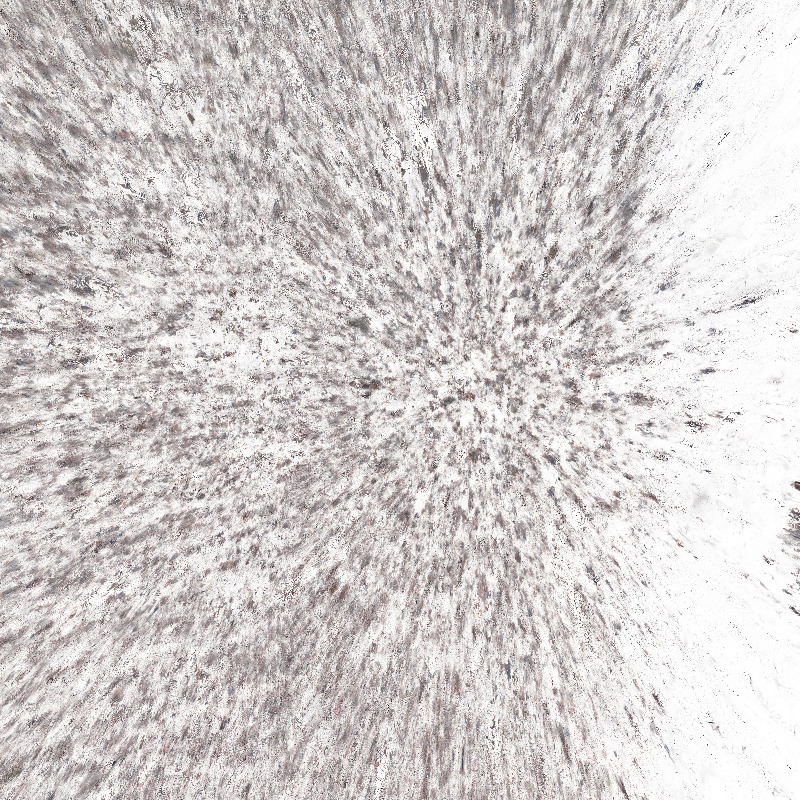} \\[1pt]
        \rotatebox{90}{\small\;\;Materials} &
        \includegraphics[width=0.175\linewidth]{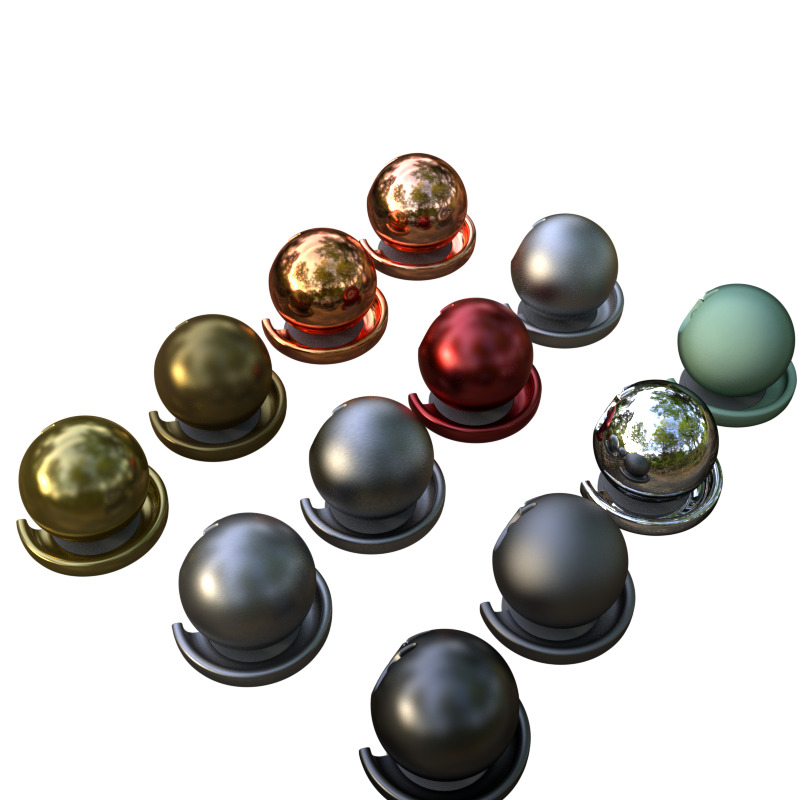} &
        \includegraphics[width=0.175\linewidth]{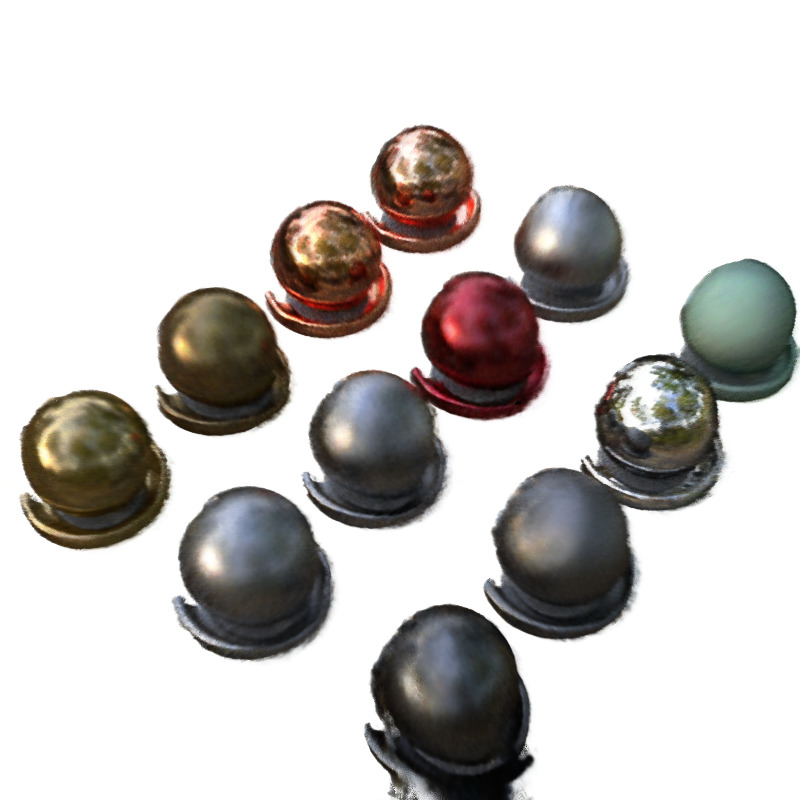} &
        \includegraphics[width=0.175\linewidth]{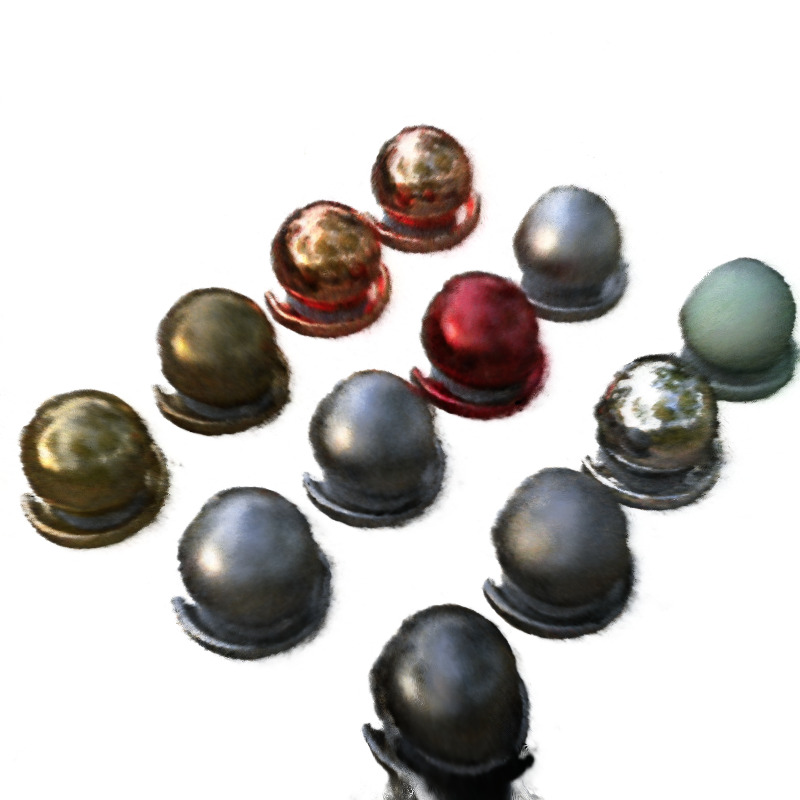} &
        \includegraphics[width=0.175\linewidth]{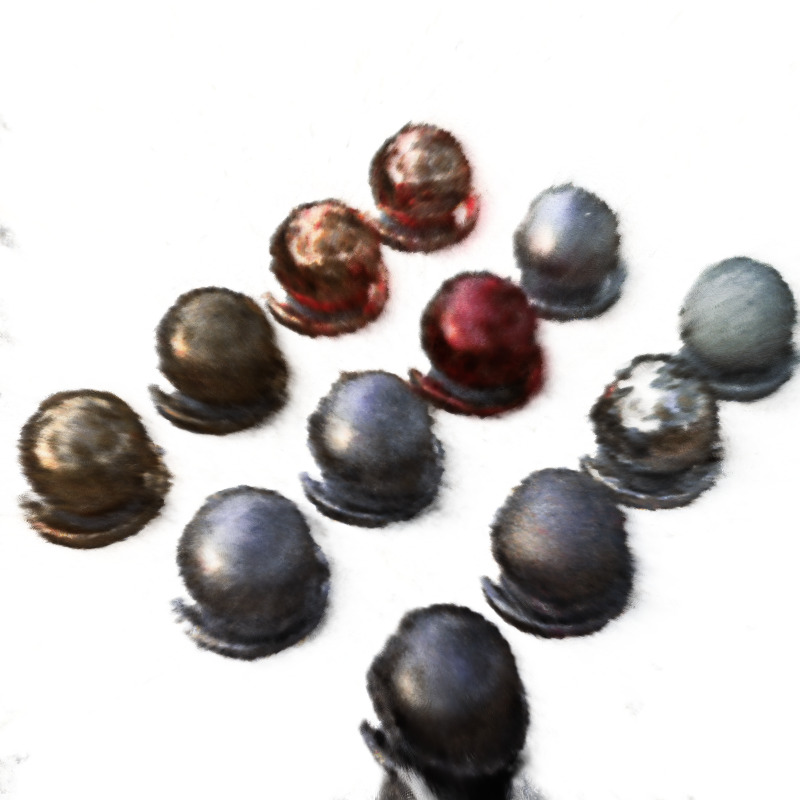} &
        \includegraphics[width=0.175\linewidth]{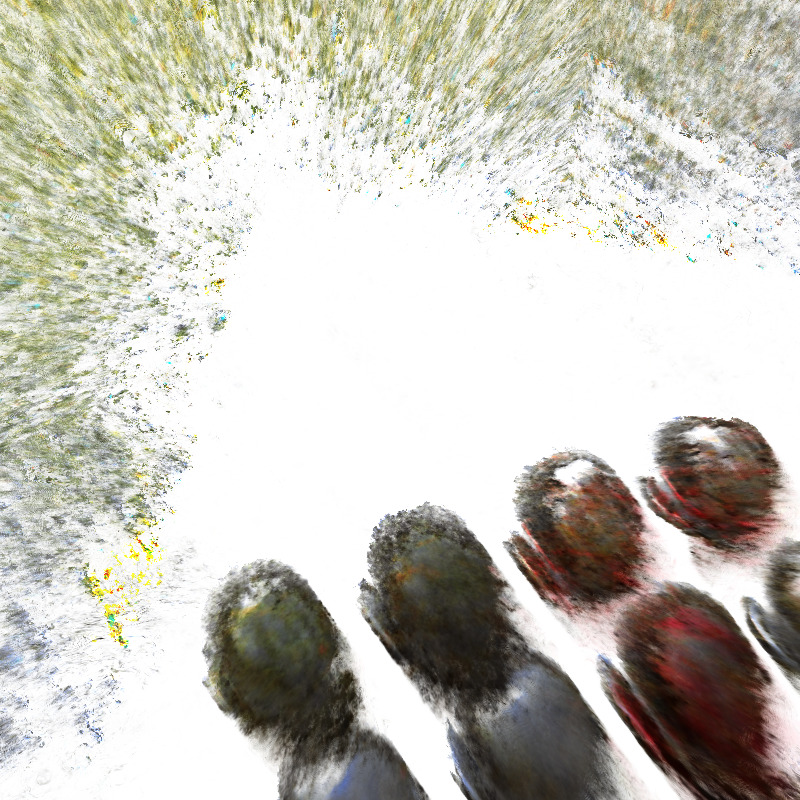} \\[1pt]
        \rotatebox{90}{\small\;\;Hotdog} &
        \includegraphics[width=0.175\linewidth]{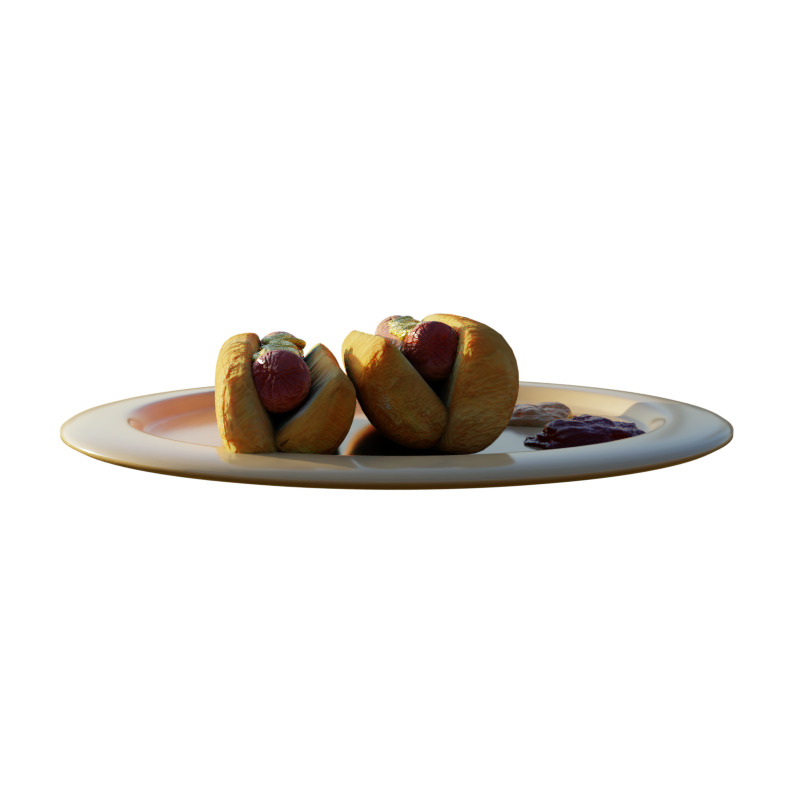} &
        \includegraphics[width=0.175\linewidth]{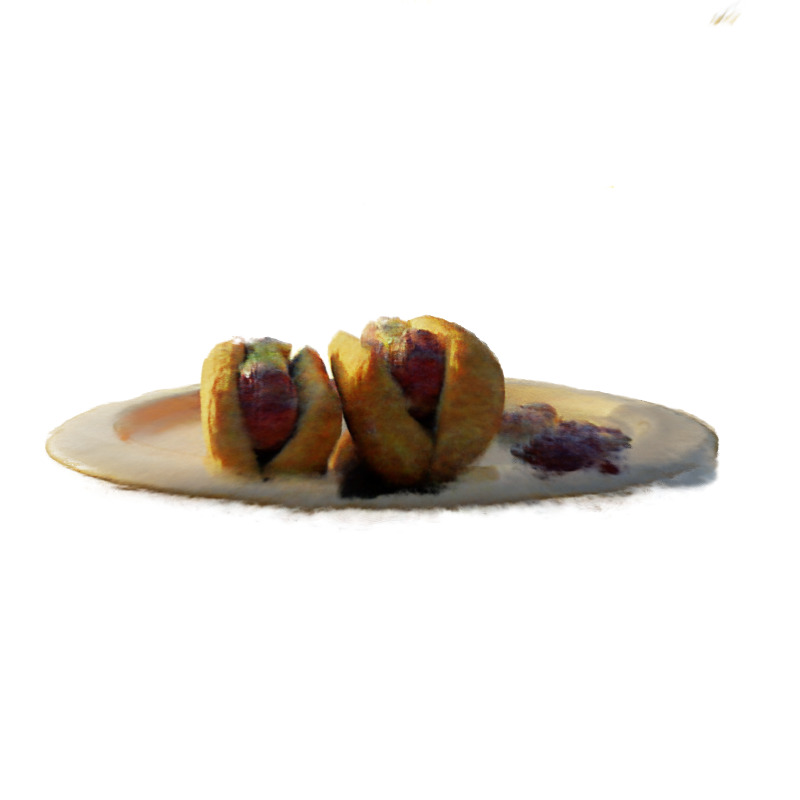} &
        \includegraphics[width=0.175\linewidth]{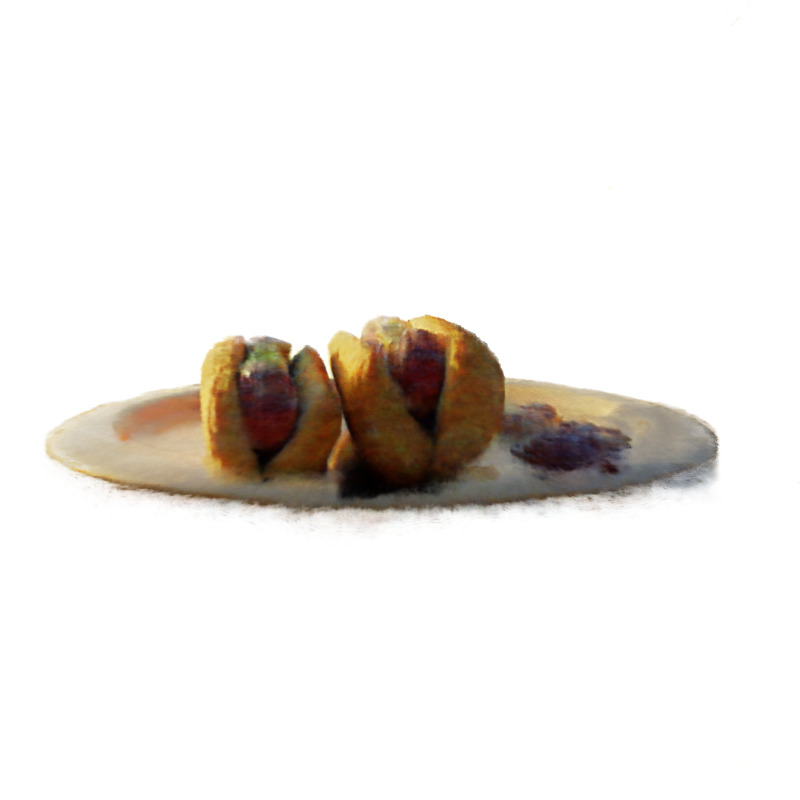} &
        \includegraphics[width=0.175\linewidth]{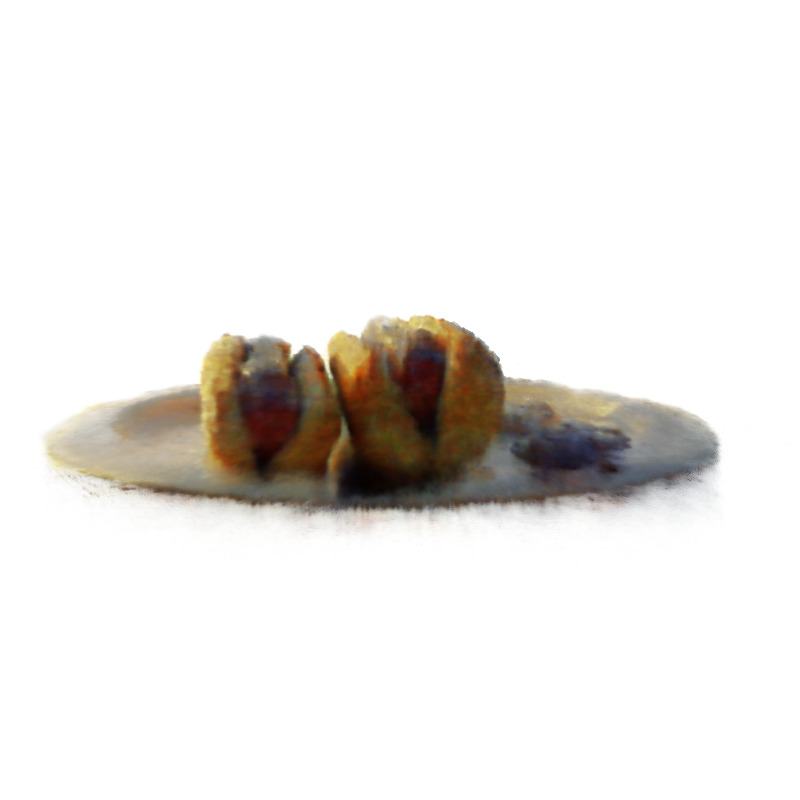} &
        \includegraphics[width=0.175\linewidth]{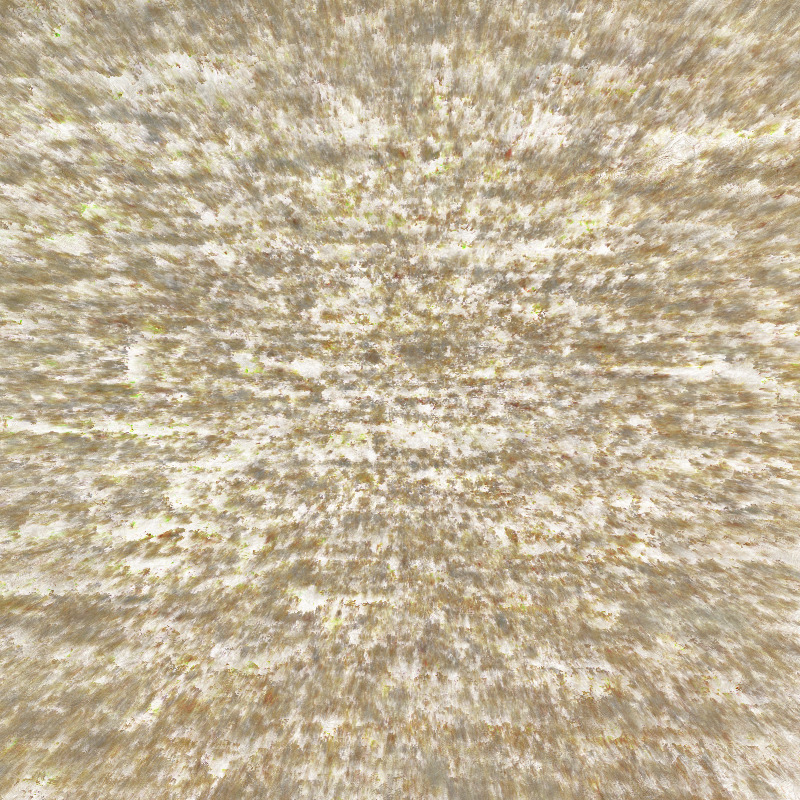} \\[1pt]
        \rotatebox{90}{\small\;\;Drums} &
        \includegraphics[width=0.175\linewidth]{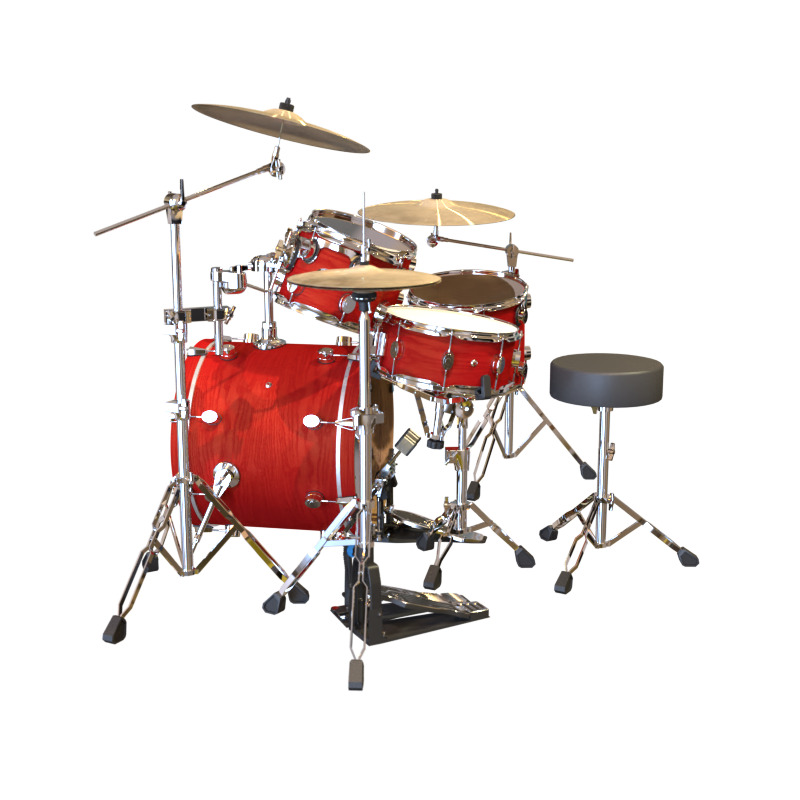} &
        \includegraphics[width=0.175\linewidth]{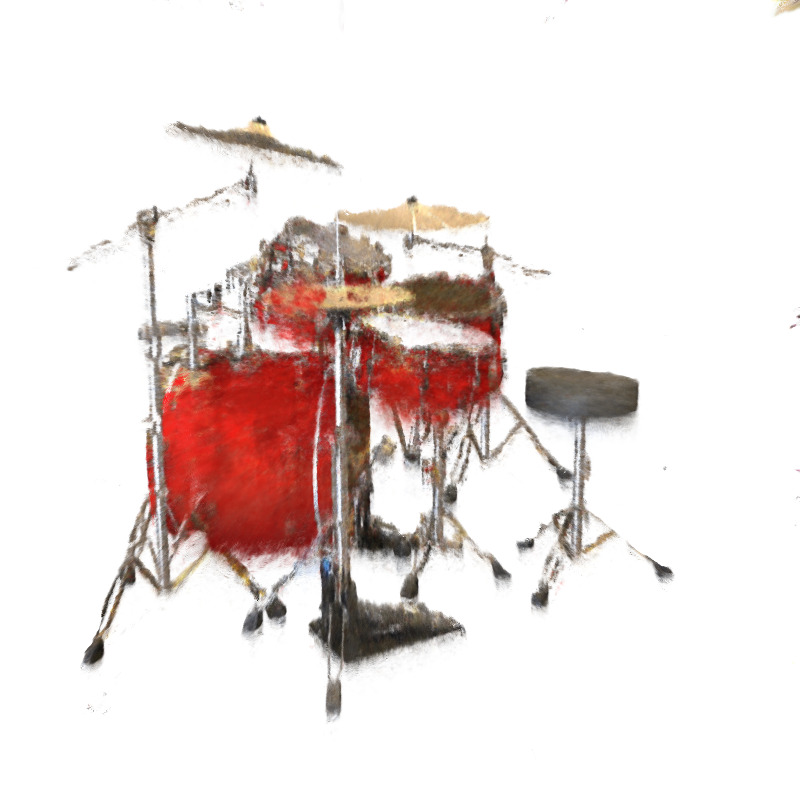} &
        \includegraphics[width=0.175\linewidth]{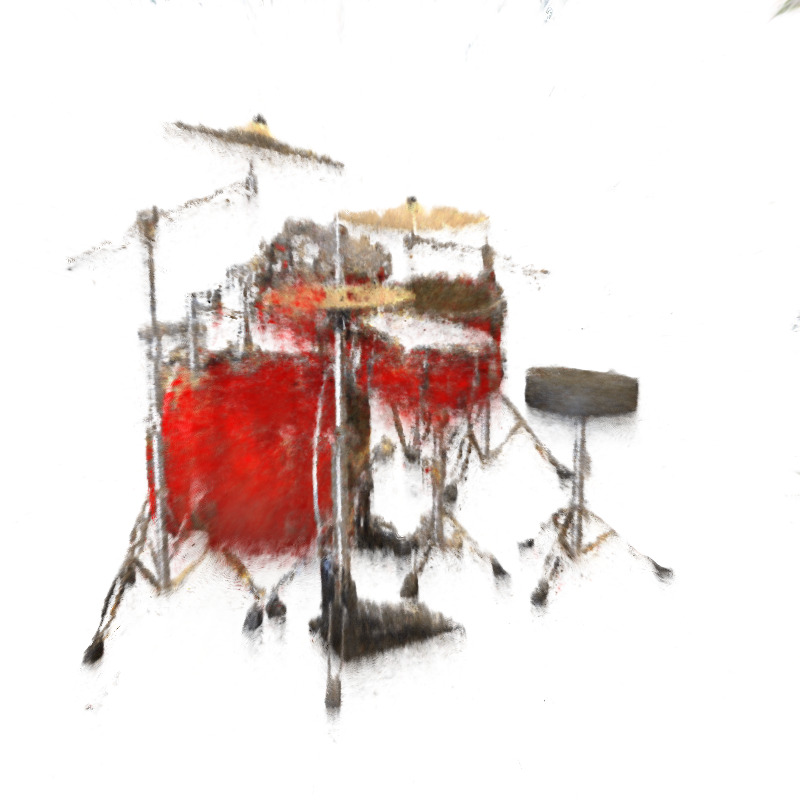} &
        \includegraphics[width=0.175\linewidth]{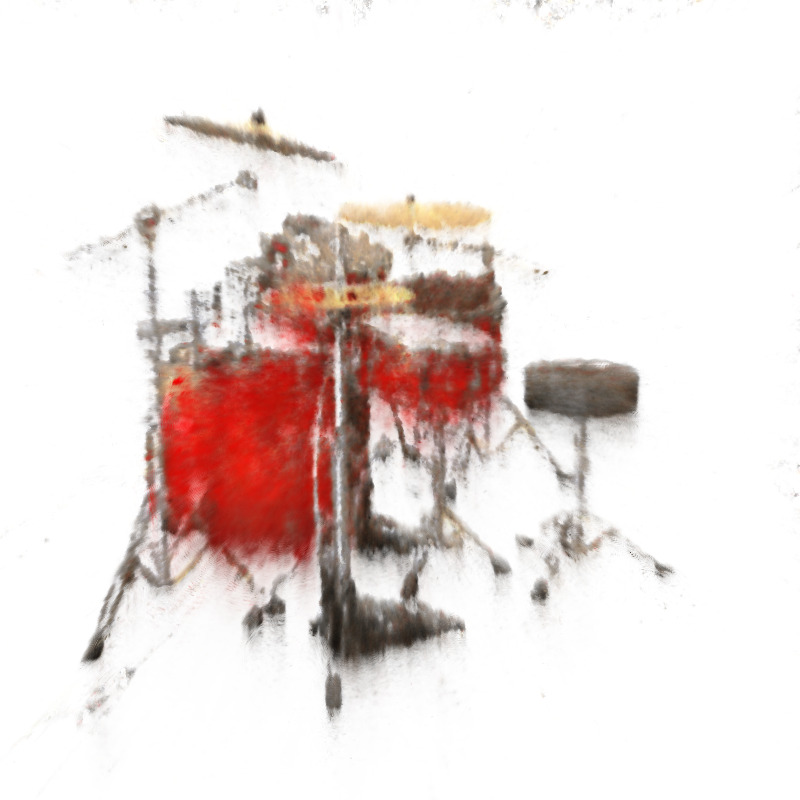} &
        \includegraphics[width=0.175\linewidth]{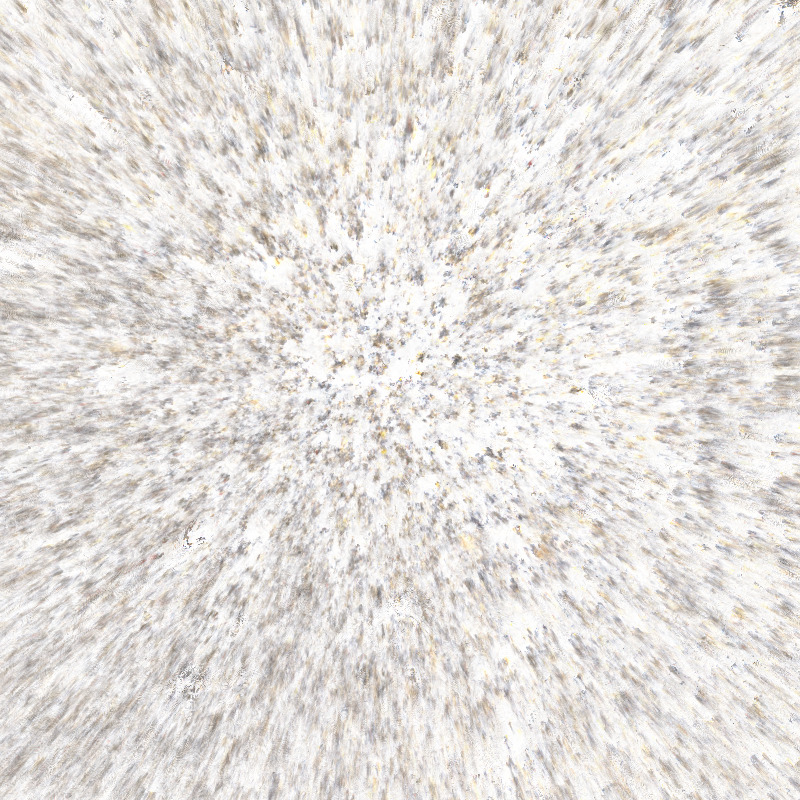} \\
    \end{tabular}
    \caption{Additional qualitative comparisons on four NeRF-Synthetic scenes under the 3-view protocol. Columns left to right: ground truth, DriftNeRF (ours), DNGaussian, FreeNeRF, NeRF.}
    \Description{A four-row by five-column grid of renderings for Ship, Materials, Hotdog, and Drums. Columns are ground truth, DriftNeRF, DNGaussian, FreeNeRF, and NeRF. The NeRF column is the haziest with the most floaters; DriftNeRF is closest to the ground truth on Hotdog and Drums, while Ship and Materials remain difficult for all methods.}
    \label{fig:qualitative_supp}
\end{figure*}

\Cref{fig:qualitative_supp} presents renderings from four additional NeRF-Synthetic scenes under the 3-view protocol.

\end{document}